%% file: naacl_ArXiv.tex
\documentclass[11pt]{article}

\usepackage[final]{acl}

\usepackage{times}
\usepackage{latexsym}

\usepackage[T1]{fontenc}

\usepackage[utf8]{inputenc}

\usepackage{microtype}

\usepackage{inconsolata}

\usepackage{graphicx}

\usepackage{booktabs}

\usepackage{colortbl}
\definecolor{LCHeaderGray}{HTML}{E6E6E6}
\usepackage{amsmath}
\usepackage{amssymb}
\usepackage{float}
\usepackage{stfloats}
\usepackage{placeins}

\usepackage{tabularx}
\usepackage{enumitem}

\definecolor{LCVenueGray}{HTML}{888888}

\usepackage[colaction]{multicol}
\usepackage{needspace}
\usepackage{capt-of}

\title{LeanContext: Disentangling Mathematical Resources\\from Local Formal Context in Lean Whole-Proof Generation}

\author{
 \textbf{Hang Zhang\textsuperscript{1}\thanks{Equal contribution.}},
 \textbf{Qihao Ye\textsuperscript{2}\footnotemark[1]},
 \textbf{Litong Liu\textsuperscript{3}\footnotemark[1]},
 \textbf{Yunbei Zhang\textsuperscript{4}},
 \textbf{Yugyung Lee\textsuperscript{1}},
\\
 \textbf{Yue Zhou\textsuperscript{5}},
 \textbf{Bowen Liu\textsuperscript{1}},
 \textbf{Emily First\textsuperscript{6}},
 \textbf{Cheng Han\textsuperscript{1}\footnotemark[2]},
 \textbf{Qiao Zhuang\textsuperscript{1}\thanks{Corresponding authors}}
\\
\\
 \textsuperscript{1}University of Missouri-Kansas City,
 \textsuperscript{2}Columbia University,
 \textsuperscript{3}Monash University Malaysia,
\\
 \textsuperscript{4}Tulane University,
 \textsuperscript{5}LLM Core and Agentic AI, Capital One,
 \textsuperscript{6}Rutgers University
 }

\begin{document}
\raggedbottom
\setlength{\intextsep}{6pt plus 2pt minus 2pt}
\setlength{\dbltextfloatsep}{8pt plus 2pt minus 2pt}
\maketitle
\begin{abstract}
Neural theorem proving increasingly uses retrieved premises and project context to support proof generation. Evaluating these resources requires measuring the advantage of target-selected declarations over comparable alternatives and how this advantage depends on training. We introduce \textsc{LeanContext}, a benchmark of 129,826 Lean whole-proof generation problems, to investigate these questions. We group selected complete preceding mathematical declarations as Mathematical Context (MC) and the remaining selected local source commands as Environment Context (EC). Four aligned input views support crossed training and inference evaluations with shared targets and original Lean verification environments. For both diffusion- and autoregressive-based models, selected declarations outperform comparable alternatives by a larger margin when training includes MC. Final-input audits identify omitted declarations referenced by the target. Prioritizing these dependencies improves Pass@32 by 1.40\% and 1.70\% for LLaDA and Llama, respectively, on 2,000 targets. The selection experiment operates without access to target reference proofs and holds checkpoints, EC, and inference budgets constant. The code and dataset are available at \url{https://github.com/haaaardcvZ/LeanContext}.
\end{abstract}

\section{Introduction}
\label{sec:introduction}

Neural theorem proving increasingly uses retrieved premises and project source context to generate proofs~\citep{yang2023leandojo,hu2025minictx}. This development reflects the cumulative nature of formal mathematics, where definitions introduce objects and subsequent lemmas establish properties that later proofs reuse~\citep{mathlib2020}. A proof about a newly defined object may combine its definition, an earlier lemma, and an instance required to apply that lemma. Supporting such proof construction requires models to use the mathematical declarations supplied by the current project.

Existing work has examined several forms of information that support this process. LeanDojo studies retrieval-augmented proving with accessible premises~\citep{yang2023leandojo}. Passport examines identifier representations, while Rango retrieves relevant premises and proof examples~\citep{sanchezstern2023passport,thompson2025rango}. miniCTX extends evaluation to project source context and analyzes the contributions of environment information, definitions, lemma statements, proofs, and comments~\citep{hu2025minictx}. These studies establish that different information components can help proof generation. We examine the advantage of target-selected declarations over comparable replacements and how training with declarations changes this advantage when accompanying source commands are held fixed.

Answering this question calls for a common unit of comparison across mathematical resources. Definitions, lemmas, structures, classes, and instances serve different functions, but subsequent proofs can reuse them as declarations. Treating a \emph{complete declaration} as the resource unit allows its identity to be tracked through selection, replacement, and dependency analysis. The comparison must also preserve other visible source commands, such as \texttt{variable}, \texttt{namespace}, and \texttt{notation}, when the declaration set changes. This motivates organizing declaration resources separately from the commands that accompany them. Declarations matched for kind, scope, and length can differ in their usefulness for proving a given target. Comparing these alternatives with the accompanying source commands held fixed makes the advantage of the selected content measurable. Verified proof success provides a common criterion for evaluating this advantage across different valid proof strategies.

\begin{figure*}[!tp]
    \centering
    \includegraphics[width=\textwidth]{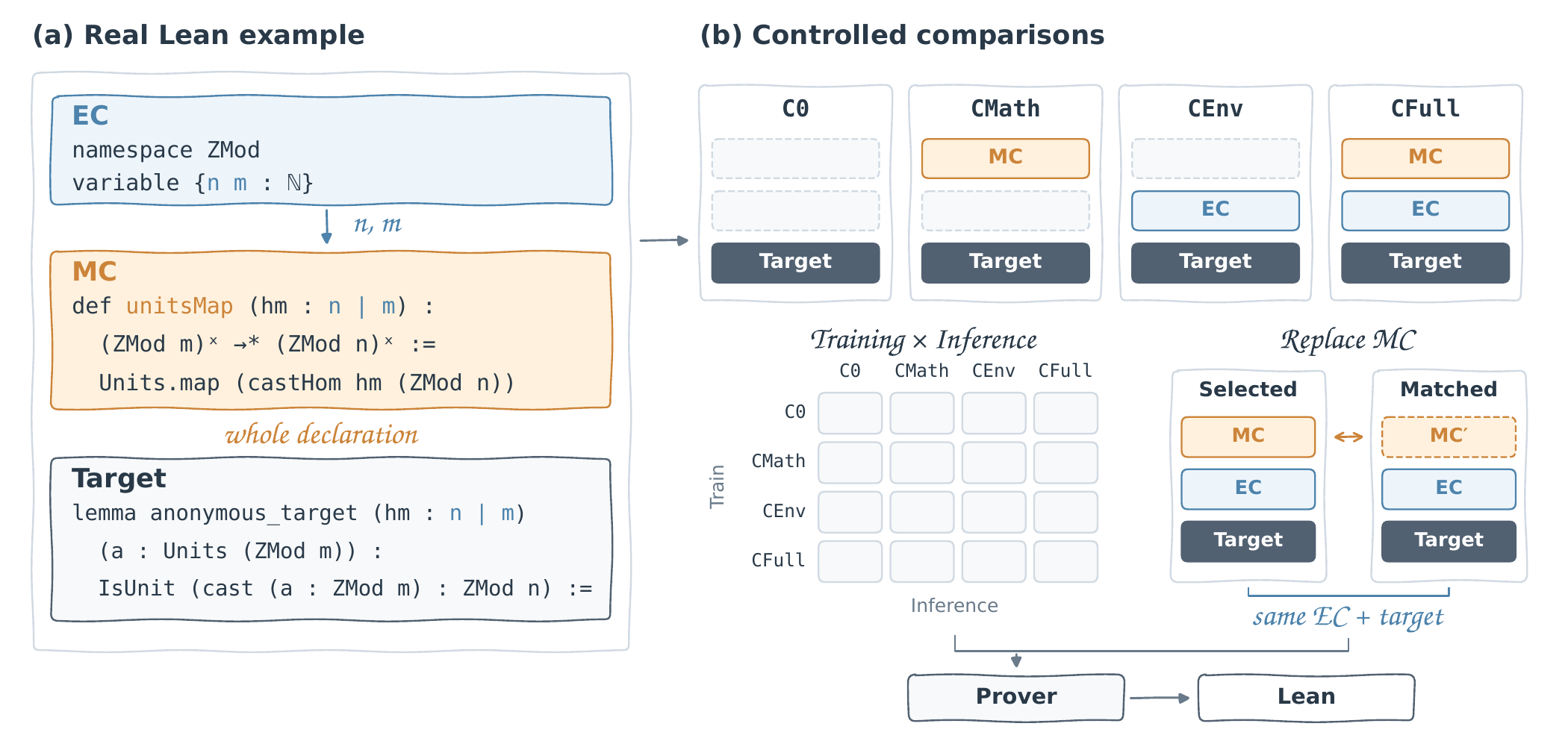}
    \caption{\textbf{Overview of \textsc{LeanContext}.} (Left) A complete mathematical declaration (MC) and accompanying source commands (EC) precede a proof target. (Right) Four aligned views support crossed training, inference evaluations, and matched replacement. Input comparisons retain the target and original Lean verification environment. Matched replacement also keeps EC unchanged.}
    \label{fig:leancontext_overview}
\end{figure*}

We introduce \textsc{LeanContext}, a benchmark of 129,826 Lean whole-proof generation problems from a fixed Mathlib4 snapshot. Mathematical Context (MC) comprises selected complete preceding mathematical declarations, while Environment Context (EC) comprises the remaining selected local source commands. 
The MC/EC partition supports comparisons of selected mathematical declarations with matched alternatives and enables us to examine how their advantage changes across training conditions.
Four aligned input views supply neither category, MC alone, EC alone, or both (Full). They support crossed training and inference evaluations, as well as declaration replacements with EC held constant. Generated proofs are verified at the original target location in the complete Lean environment. Figure~\ref{fig:leancontext_overview} illustrates the context partition and these comparisons.

To examine training-dependent declaration benefits under both diffusion and autoregressive generation, we evaluate LLaDA~\citep{nie2025large} and Llama~\citep{llama2024herd}, respectively. With checkpoints trained on both context categories, adding MC to EC improves Pass@32 by 7.1\% and 8.5\%, respectively. On the same 11,064 test targets, the corresponding gains are 0.4\% and 0.8\% after EC-only training. Comparisons with alternatives matched for kind, scope, and length further show a larger advantage for selected declarations when training includes MC. Dependency audits then identify declarations omitted from final inputs. Prioritizing these dependencies without accessing target reference proofs improves verified success under fixed checkpoints, EC, and budgets, while revealing resource tradeoffs.

Our main contributions are as follows:
\begin{itemize}[noitemsep, topsep=3pt, leftmargin=*]
    \item \textbf{A benchmark for controlled context comparisons.} We provide four aligned input views for 129,826 Lean proof problems, supporting comparisons of declaration visibility under fixed targets and verification environments.
    \item \textbf{Evidence on training-dependent declaration benefits.} Crossed evaluations and matched replacements show larger declaration gains and selected-content advantages after training with MC than after EC-only training in both evaluated backbones.
    \item \textbf{Dependency diagnostics that guide selection.} We audit declaration visibility in final inputs and use these diagnostics to test a dependency-priority policy that improves verified proof success.
\end{itemize}

\section{Related Work}
\label{sec:related_work}

\paragraph{Premises and Project Context.}
Premise retrieval supplies mathematical resources beyond the current proof state~\citep{alemi2016deepmath,mikula2024magnushammer}. LeanDojo provides accessible-premise annotations and supports retrieval-augmented proving in Lean~\citep{yang2023leandojo}. Research has also examined how contextual information is represented and presented. Passport studies identifier categories and representations~\citep{sanchezstern2023passport}, while Rango retrieves lemma statements and proof examples~\citep{thompson2025rango}. miniCTX extends evaluation to preceding project context and analyzes environment information, definitions, lemma statements, lemma proofs, and comments~\citep{hu2025minictx}. These studies broaden context beyond retrieved premises and examine the contributions of different information components.

\paragraph{Training with Context.}
The use of supplied resources also depends on training. miniCTX investigates file-tuning on preceding source context, and Rango evaluates training variants with different retrieval components~\citep{hu2025minictx,thompson2025rango}. AutoReal combines reasoning-annotated proof training with context augmentation and compares base and fine-tuned models with and without context~\citep{zhang2026autoreal}. These comparisons examine both the information supplied to a prover and the training that prepares it to use that information.

\paragraph{Evaluating Supplied Resources.}
Recent evaluations examine the value of supplied content more directly. FormalML compares absent premises, ground-truth premises, and candidate sets containing relevant premises and distractors~\citep{yang2026formalml}. VeriSoftBench compares curated proof dependencies with broader repository context while retaining those dependencies during budget handling~\citep{xin2026verisoftbench}. LeanSearch v2 evaluates retrieval through downstream proof success under a shared prover and reflection workflow~\citep{gao2026leansearchv2}.

Our proposed \textsc{LeanContext} focuses on whether declarations selected for a target outperform comparable replacements and how this advantage depends on training. It uses complete declarations as the resource unit and holds accompanying source commands fixed in matched comparisons. Crossed training and inference views assess training dependence, while final-input dependency audits guide a tested selection change. Additional related work is discussed in Appendix~\ref{sec:additional_related_work}.

\section{LeanContext: A Context-disentangled Lean Benchmark}
\label{sec:benchmark}

Figure~\ref{fig:leancontext_overview} illustrates the declaration-level context partition, aligned input views, and controlled comparisons used in \textsc{LeanContext}.

\subsection{Task and Context Partition}
\label{sec:resource_representation}

We evaluate whole-proof generation~\citep{first2023baldur}: a model receives a target statement and visible preceding source commands and generates a complete proof without intermediate Lean feedback. Each candidate is checked at the original target location with Lean~\citep{demoura2021lean4} in the complete original environment. The input views therefore vary model visibility while preserving verification conditions.

We organize context by complete source commands. MC groups mathematical declarations, including definitions, lemmas, structures, classes, and instances. Their statements and bodies are retained when present, including preceding lemma proofs. EC groups the remaining selected commands, such as \texttt{variable}, \texttt{namespace}, \texttt{open}, and \texttt{notation}. The partition follows source command kinds, whose mathematical and scope-related roles can overlap. Matched comparisons vary declaration resources with EC held constant. The complete command-kind partition is provided in Appendix~\ref{sec:appendix_command_classification}.

\subsection{Dataset and Aligned Views}
\label{sec:dataset_construction}

Our benchmark contains 129,826 theorem, lemma, and example targets from 4,469 files in a fixed Mathlib4 snapshot~\citep{mathlib2020}. Records retain reference proofs and selected preceding commands from the same file, with source identities and spans. The original selection can inspect target reference proofs, which are excluded from evaluation inputs. Source files and normalized target statements are disjoint across splits (Table~\ref{tab:dataset_splits}). Extraction, filtering, and complete dataset statistics are detailed in Appendix~\ref{sec:appendix_construction}.

\begin{table}[H]
    \centering
    \small
    \setlength{\tabcolsep}{6pt}
    \setlength{\aboverulesep}{0pt}
    \setlength{\belowrulesep}{0pt}
    \setlength{\extrarowheight}{1pt}
    \begin{tabular}{lrr}
        \toprule
        \rowcolor{LCHeaderGray}
        Split & Targets & Both \\
        \midrule
        Train  & 90,879  & 51,628 \\
        Valid. & 19,473  & 11,063 \\
        Test   & 19,474  & 11,064 \\
        \midrule
        Total  & 129,826 & 73,755 \\
        \bottomrule
    \end{tabular}
    \caption{\textbf{Dataset statistics.}
    Both counts targets with nonempty selected MC
    and EC before model-specific budget handling.}
    \label{tab:dataset_splits}
\end{table}

Each target has four views: \texttt{C0} supplies neither category, \texttt{CMath} supplies MC, \texttt{CEnv} supplies EC, and \texttt{CFull} supplies both (Figure~\ref{fig:leancontext_overview}). We first fit complete commands into \texttt{CFull} under the model's 1,024-token input limit, including the target and special tokens. Other views hide categories without refilling the freed space. Paired inputs thus retain identical targets, special tokens, and commands for the category held fixed. The primary evaluation uses the 11,064 test targets with both categories available before budget handling. Dataset records are illustrated in Appendix~\ref{sec:appendix_dataset_records}, and serialization and budget handling are detailed in Appendix~\ref{sec:appendix_input_pipeline}.

\subsection{Controlled Declaration Comparisons}
\label{sec:resource_comparisons}

For each backbone, four training runs differ only in input view, sharing targets, reference-proof supervision, initialization, and optimization settings. Each checkpoint is evaluated on all four inference views. Let $S_\theta(v)$ denote Pass@32 on a fixed target set. Conditional category gains are
\begin{align}
    \Delta_{M\mid E}(\theta)
    &= S_\theta(\mathrm{CFull})-S_\theta(\mathrm{CEnv}),
    \label{eq:mc_gain}\\
    \Delta_{E\mid M}(\theta)
    &= S_\theta(\mathrm{CFull})-S_\theta(\mathrm{CMath}).
    \label{eq:ec_gain}
\end{align}
Each contrast uses a single checkpoint. Comparing gains across training views examines their dependence on training exposure.

Matched replacement tests the value of selected content on 2,000 matching-eligible targets, without selecting by generation outcomes. Alternatives match declaration kind, compatible scope, and token length within $\pm10\%$ under both tokenizers. They exclude identified target-type and reference-proof dependencies, making this comparison reference-assisted. With target, EC, checkpoint, verifier, and budgets fixed, the selected-content advantage is
\begin{equation}
    R_\theta
    = S_\theta(E+M_{\mathrm{sel}})
    - S_\theta(E+M_{\mathrm{rep}}),
    \label{eq:selected_content}
\end{equation}
where $M_{\mathrm{sel}}$ and $M_{\mathrm{rep}}$ are selected and replacement MC, and $E$ is the shared EC. We repeat this comparison under \texttt{CMath}, \texttt{CEnv}, and \texttt{CFull} training. Subset construction and matching rules are detailed in Appendix~\ref{sec:appendix_matching_protocol}.

We also trace selected-MC reuse in verified proofs and classify failed candidates using a fixed precedence over extraction status and Lean diagnostics. Reuse criteria and candidate-classification rules are detailed in Appendices~\ref{sec:appendix_verified_reuse} and~\ref{sec:appendix_candidate_outcomes}. We further examine representation responses across views. Centered linear CKA~\citep{kornblith2019similarity} compares token-averaged hidden states from the central proof region (25\% to 75\%) on 4,000 shared test targets under fixed reference-proof conditioning. We observe blocks 8, 16, 24, and 32 and report $1-\mathrm{CKA}$. Model-specific readouts and pooling are detailed in Appendix~\ref{sec:appendix_cka}.

\subsection{Dependency Diagnostics and Selection}
\label{sec:dependency_selection}

To audit actual declaration visibility, we map constants in each target's elaborated type to eligible preceding local MC commands, deduplicating command identities to form $D_i^T$. This set is independent of selected MC and the candidate pool. Let $V_i^M$ contain complete MC commands visible in the final serialized input. For nonempty $D_i^T$ with resolved mapping and visibility,
\begin{equation}
    \operatorname{Coverage}_T(i)
    = |D_i^T\cap V_i^M|/|D_i^T|.
    \label{eq:target_coverage}
\end{equation}
We distinguish zero, partial, and complete coverage. Targets with no eligible dependencies or unresolved evidence are reported separately.

Target-type dependencies identify declarations referenced by the proposition to prove. We separately audit additional dependencies found in the reference proof. Selection priority uses only target-type dependencies. Additional reference-proof dependencies are audited afterward to inspect resource tradeoffs. These resources compete for the same input budget.

Selection controls use a fixed 2,000-target subset distinct from the matched subset. Rebuilding context without target reference proofs or proof-derived features can change both MC and EC. A separate dependency-priority comparison uses the same targets with a common candidate pool built independently of the original selector, without reference proofs or proof-derived features.

The baseline ranks declarations by lexical overlap with the target. Dependency priority places eligible target dependencies first and preserves baseline order within each group. Both policies freeze the same EC and retain identical checkpoints, verification conditions, and inference budgets. Rules are fixed on validation data, and all 2,000 targets remain in the success denominator. Additional reference-proof dependencies are audited only after selection and generation outcomes are fixed, to examine resource tradeoffs. Full matching, mapping, and selection procedures are provided in Appendix~\ref{sec:appendix_dependencies}.

\section{Experiments and Analysis}
\label{sec:experiments}

We examine training-dependent declaration benefits, then connect selected-content advantages to verified proof behavior and representation responses. Dependency diagnostics then test whether the benchmark can guide a useful selection change.

\subsection{Experimental Setup}
\label{sec:experimental_setup}

Our primary backbones cover diffusion and autoregressive generation: LLaDA-8B-Base~\citep{nie2025large} and Llama-3.1-8B~\citep{llama2024herd}, respectively. Each is adapted through full-parameter SFT under the four training views in Section~\ref{sec:resource_comparisons}. Checkpoints are selected by validation loss. The main crossing uses 11,064 test targets with both categories available before budget handling. Matched replacement and selection use distinct 2,000-target subsets, and coverage is evaluated separately on 11,064 targets.

Inputs and per-attempt outputs are each set to 1,024 tokens. We generate 32 candidates per target at temperature 0.6 without intermediate feedback and verify each at its original Lean location, with a 60-second limit. Sampled Pass@$k$ is estimated from these outcomes~\citep{chen2021evaluating}. All targets remain in the success denominator, including those without submitted inputs. Scores and absolute score differences are reported in \%. Full configurations and expanded results are provided in Appendix~\ref{sec:appendix_experiments}.

\begin{table*}[!t]
    \centering
    \small
    \setlength{\tabcolsep}{5pt}
    \setlength{\aboverulesep}{0pt}
    \setlength{\belowrulesep}{0pt}
    \setlength{\extrarowheight}{1pt}
    \begin{tabular}{llrrrrrr}
        \toprule
        \rowcolor{LCHeaderGray}
        \multicolumn{2}{l}{Checkpoint}
        & \multicolumn{4}{c}{Inference view / Pass@32 (\%)}
        & \multicolumn{2}{c}{Conditional gain (\%)} \\
        \cmidrule(lr){1-2}
        \cmidrule(lr){3-6}
        \cmidrule(lr){7-8}
        \rowcolor{LCHeaderGray}
        Model & Training
        & \texttt{C0} & \texttt{CMath}
        & \texttt{CEnv} & \texttt{CFull}
        & $\Delta_{M\mid E}$ & $\Delta_{E\mid M}$ \\
        \midrule
        LLaDA~{\color{LCVenueGray}\scriptsize NeurIPS'25}
            & \texttt{C0}
            & 38.6 & 39.1 & 37.8 & 38.3 & +0.5 & -0.8 \\
            & \texttt{CMath}
            & 34.1 & 43.9 & 35.8 & 44.3 & +8.5 & +0.4 \\
            & \texttt{CEnv}
            & 34.7 & 35.3 & 44.8 & 45.2
            & \textbf{+0.4} & +9.9 \\
            & \texttt{CFull}
            & 31.2 & 40.1 & 41.6 & 48.7
            & \textbf{+7.1} & +8.6 \\
        \midrule
        Llama~{\color{LCVenueGray}\scriptsize arXiv'24}
            & \texttt{C0}
            & 39.8 & 40.6 & 39.2 & 40.3 & +1.1 & -0.3 \\
            & \texttt{CMath}
            & 34.8 & 44.3 & 37.6 & 44.9 & +7.3 & +0.6 \\
            & \texttt{CEnv}
            & 34.5 & 37.3 & 44.9 & 45.7
            & \textbf{+0.8} & +8.4 \\
            & \texttt{CFull}
            & 32.8 & 41.8 & 41.1 & 49.6
            & \textbf{+8.5} & +7.8 \\
        \bottomrule
    \end{tabular}
    \caption{\textbf{Crossed training and inference views on 11,064 targets.} Inference columns report Pass@32. Each row corresponds to one checkpoint. Conditional gains follow Eqs.~\eqref{eq:mc_gain} and~\eqref{eq:ec_gain}. Bold highlights the MC-gain comparison between
    \texttt{CEnv} and \texttt{CFull} training.}
    \label{tab:training_context_crossing}
\end{table*}

\subsection{Training-Dependent Declaration Benefits}
\label{sec:training_context_results}

Checkpoints trained with MC obtain larger gains when MC is supplied at inference (Table~\ref{tab:training_context_crossing}). With EC held fixed, adding MC improves Pass@32 by 7.1\% and 8.5\% for \texttt{CFull}-trained LLaDA and Llama, compared with 0.4\% and 0.8\% after \texttt{CEnv} training. The \texttt{CMath}-trained checkpoints show the same pattern.

The crossing also compares training conditions under a common inference view (Table~\ref{tab:training_context_crossing}). Under \texttt{CFull}, \texttt{CFull}-trained checkpoints outperform \texttt{CEnv}-trained checkpoints by 3.5\% for LLaDA and 3.9\% for Llama. Under \texttt{CEnv}, the same \texttt{CFull}-trained checkpoints score 3.2\% and 3.8\% below their \texttt{CEnv}-trained counterparts, respectively. These contrasts show that the observed advantage of combined-context training depends on the resources visible at inference.

EC gains show a complementary pattern. Adding EC to \texttt{CMath} yields 8.6\% and 7.8\% after \texttt{CFull} training, versus 0.4\% and 0.6\% after \texttt{CMath} training. Both categories provide additional value after combined-context training. The crossing also captures adaptation to the input configuration: under \texttt{C0}, \texttt{CFull}-trained checkpoints score 7.4\% and 7.0\% below their \texttt{C0}-trained counterparts.

The MC benefit persists across sampling budgets (Table~\ref{tab:sampling_budget_results}), increasing from 1.0\%/1.3\% at Pass@1 to 7.1\%/8.5\% at Pass@32. Additional adaptation and temperature results, together with evaluations using the same view for training and inference, are reported in Appendix~\ref{sec:appendix_experiments}.

\begin{table}[H]
    \centering
    \small
    \setlength{\tabcolsep}{2.5pt}
    \setlength{\aboverulesep}{0pt}
    \setlength{\belowrulesep}{0pt}
    \setlength{\extrarowheight}{1pt}
    \begin{tabular}{llrrrr}
        \toprule
        \rowcolor{LCHeaderGray}
        Model & View & @1 & @8 & @16 & @32 \\
        \midrule
        LLaDA~{\color{LCVenueGray}\scriptsize NeurIPS'25}
            & \texttt{CEnv}
            & 5.9 & 23.2 & 32.3 & 41.6 \\
            & \texttt{CFull}
            & \textbf{6.9} & \textbf{27.1}
            & \textbf{37.8} & \textbf{48.7} \\
        \midrule
        Llama~{\color{LCVenueGray}\scriptsize arXiv'24}
            & \texttt{CEnv}
            & 5.9 & 23.1 & 32.0 & 41.1 \\
            & \texttt{CFull}
            & \textbf{7.2} & \textbf{27.9}
            & \textbf{38.6} & \textbf{49.6} \\
        \bottomrule
    \end{tabular}
    \caption{\textbf{Pass@$k$ on the main 11,064 targets
    for \texttt{CFull}-trained checkpoints}, estimated
    from the same 32 candidate outcomes.
    Bold marks the higher score within each model
    and sampling budget.}
    \label{tab:sampling_budget_results}
\end{table}

We then comprehensively evaluate released Lean provers without further \textsc{LeanContext} training: DeepSeek-Prover-V1.5-SFT and -RL~\citep{xin2025deepseekprover15}, DeepSeek-Prover-V2-7B~\citep{ren2025deepseekprover2}, Goedel-Prover-SFT~\citep{lin2025goedelprover}, and TheoremLlama~\citep{wang2024theoremllama}. Table~\ref{tab:released_context_results} keeps each checkpoint fixed across the four views. Conditional MC gains range from 1.6\% to 5.1\%, and conditional EC gains range from 2.3\% to 4.5\%. These results extend the context comparisons to existing Lean provers. The preceding training and inference crossing tests how context gains depend on training.

\begin{table}[!tbp]
    \centering
    \small
    \setlength{\tabcolsep}{0.8pt}
    \setlength{\aboverulesep}{0pt}
    \setlength{\belowrulesep}{0pt}
    \setlength{\extrarowheight}{1pt}
    \begin{tabular}{lrrrr}
        \toprule
        \rowcolor{LCHeaderGray}
        Model & \texttt{C0} & \texttt{CMath}
        & \texttt{CEnv} & \texttt{CFull} \\
        \midrule
        DeepSeek-Prover-V1.5-SFT~{\color{LCVenueGray}\scriptsize ICLR'25}
            & 20.7 & 26.4 & 25.8 & \textbf{30.9} \\
        DeepSeek-Prover-V1.5-RL~{\color{LCVenueGray}\scriptsize ICLR'25}
            & 22.9 & 28.1 & 27.7 & \textbf{32.6} \\
        DeepSeek-Prover-V2-7B~{\color{LCVenueGray}\scriptsize arXiv'25}
            & 29.4 & 35.6 & 34.8 & \textbf{38.9} \\
        Goedel-Prover-SFT~{\color{LCVenueGray}\scriptsize COLM'25}
            & 23.6 & 29.1 & 27.4 & \textbf{32.2} \\
        TheoremLlama~{\color{LCVenueGray}\scriptsize EMNLP'24}
            & 11.7 & 14.2 & 14.9 & \textbf{16.5} \\
        \bottomrule
    \end{tabular}
    \caption{\textbf{Released Lean provers on the same
    11,064 targets. Scores are Pass@32.}
    Each checkpoint is fixed across views, without
    additional \textsc{LeanContext} training.
    DeepSeek-Prover-V2-7B uses non-CoT generation.
    Bold marks the highest score within each checkpoint.}
    \label{tab:released_context_results}
\end{table}


\subsection{Selected Content and Proof Behavior}
\label{sec:matched_content_results}

\begin{figure*}[!b]

    \centering
     \includegraphics[width=\textwidth]{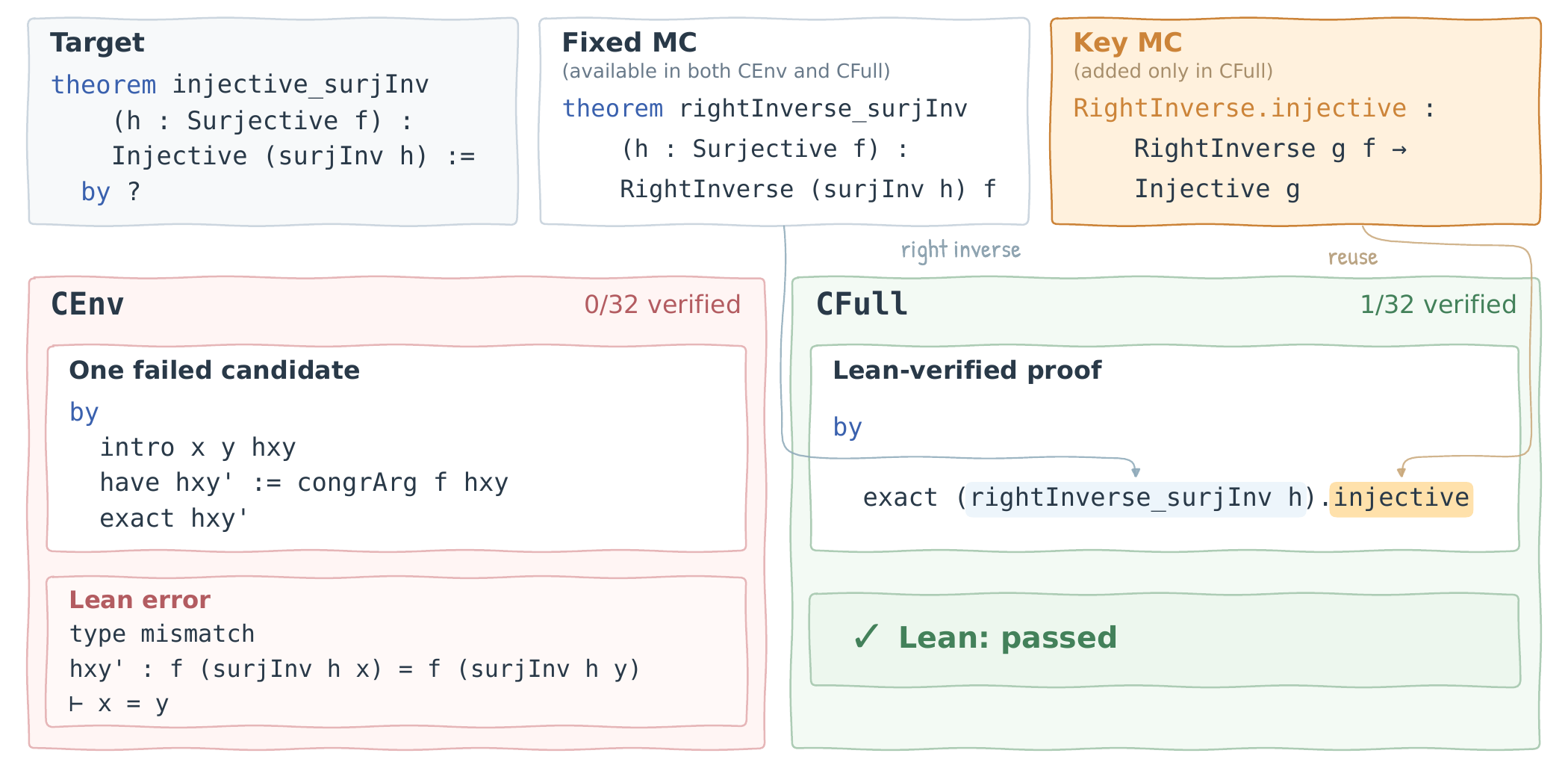}
    \caption{\textbf{A paired proof-generation case with 32 attempts per view.} Both views use the same checkpoint, EC, and verification conditions. A failed \texttt{CEnv} candidate leaves a type mismatch. A verified \texttt{CFull} candidate uses the supplied \texttt{RightInverse.injective} declaration. The right-inverse fact is available in the original Lean environment. These candidates are independently generated under the two input views.}
    \label{fig:mc_proof_case}

    \end{figure*}
    
We then test whether the training differences in Section~\ref{sec:training_context_results} can extend to the advantage of the particular declarations selected for each target.
Matched replacement compares selected MC with alternatives of the same kind, compatible scope, and similar length, retaining the same EC and checkpoint (Section~\ref{sec:resource_comparisons}). On the common 2,000-target subset, selected-content advantages are 6.0\% and 7.2\% after \texttt{CFull} training, versus 0.7\% and 0.9\% after \texttt{CEnv} training (Table~\ref{tab:matched_content_results}). The \texttt{CMath}-trained checkpoints show the same pattern. Replacement MC also improves over EC alone by 1.1\% and 1.2\% after \texttt{CFull} training. These results support assessing declaration selection together with prover training.

\begin{table}[!tbp]
    \centering
    \small
    \setlength{\tabcolsep}{1.5pt}
    \setlength{\aboverulesep}{0pt}
    \setlength{\belowrulesep}{0pt}
    \setlength{\extrarowheight}{1pt}
    \begin{tabular}{llrrrr}
        \toprule
        \rowcolor{LCHeaderGray}
        Model & Train & \texttt{CEnv}
        & Selected & Replaced & $R_\theta$ \\
        \midrule
        LLaDA~{\color{LCVenueGray}\scriptsize NeurIPS'25}
            & \texttt{CMath}
            & 35.6 & 44.1 & 37.8 & +6.3 \\
            & \texttt{CEnv}
            & 44.5 & 45.0 & 44.3 & \textbf{+0.7} \\
            & \texttt{CFull}
            & 41.3 & 48.4 & 42.4 & \textbf{+6.0} \\
        \midrule
        Llama~{\color{LCVenueGray}\scriptsize arXiv'24}
            & \texttt{CMath}
            & 37.4 & 44.8 & 39.2 & +5.6 \\
            & \texttt{CEnv}
            & 44.8 & 45.6 & 44.7 & \textbf{+0.9} \\
            & \texttt{CFull}
            & 40.9 & 49.3 & 42.1 & \textbf{+7.2} \\
        \bottomrule
    \end{tabular}
    \caption{\textbf{Matched replacement on 2,000 targets, reporting Pass@32.} Selected and Replaced combine their respective MC with identical EC. $R_\theta$ is the selected-content advantage (Eq.~\eqref{eq:selected_content}).
    Controls exclude identified target-type and
    reference-proof dependencies. Bold contrasts \texttt{CEnv} and \texttt{CFull} training.}
    \label{tab:matched_content_results}
\end{table}

Figure~\ref{fig:mc_proof_case} shows how this use appears in a concrete proof. For \texttt{Function.injective\_surjInv}, \texttt{CEnv} yields no verified proof in 32 attempts, while \texttt{CFull} yields one. The failed candidate applies \texttt{congrArg f} but leaves the required equality unresolved. The successful candidate uses \texttt{RightInverse.injective} to convert the right-inverse property of \texttt{surjInv} into injectivity. This illustrates how a supplied declaration supports proof construction under the same verification environment.

\begin{figure*}[!t]

    \centering
    \includegraphics[width=\textwidth]{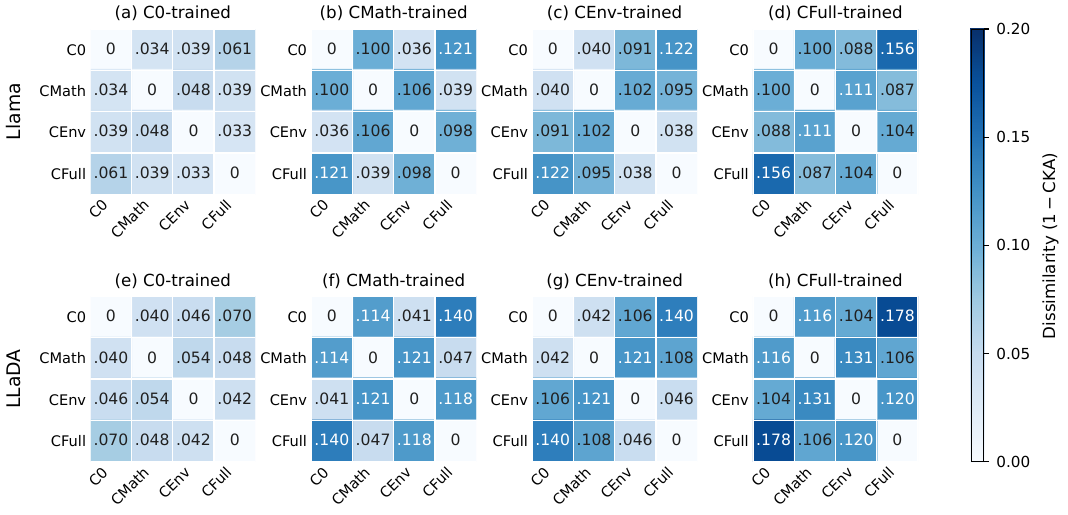}
    \caption{\textbf{Block-32 representation dissimilarity on 4,000 shared test targets under reference-proof conditioning.} Rows identify backbones, and columns identify training views. Both axes within each panel are inference views. Cells report $1-\mathrm{CKA}$ with a shared color scale. Larger values indicate lower representation similarity.}
    \label{fig:cka_heatmaps}

    \end{figure*}

We then inspect resource use in verified generations from the main crossing. At fixed \texttt{CFull}-trained checkpoints, 1,300 LLaDA targets and 1,450 Llama targets have no successful candidate under \texttt{CEnv} and at least one under \texttt{CFull}. Among them, 855 (65.77\%) and 1,005 (69.31\%) have a verified proof with recorded selected-MC reuse. These outcomes connect aggregate gains to recorded declaration use. Complete paired counts are given in Table~\ref{tab:verified_mc_reuse}. Both views retain the complete Lean environment, so recorded reuse does not establish that displaying a declaration was necessary.

We also inspect how EC changes candidate outcomes when MC is already visible. At fixed \texttt{CFull}-trained checkpoints, the \texttt{CMath}--\texttt{CFull} comparison reduces the combined share of name/scope, instance-synthesis, and other-elaboration failures from 40.9\% to 39.7\% for LLaDA and from 39.6\% to 38.4\% for Llama. Each view contains 354,048 attempts on the same 11,064 targets. These shifts characterize the observed benefit of supplying EC alongside MC. The comparison uses independently generated candidates under each view; complete outcome counts are reported in Appendix~\ref{sec:appendix_candidate_outcomes}, Table~\ref{tab:candidate_verification_outcomes}. Input-length and category-boundary sensitivity results are reported in Appendix~\ref{sec:appendix_context_sensitivity}.

MC gains are positive across all four input-length quartiles for \texttt{CFull}-trained checkpoints. The quartiles group the 11,064 targets by pre-clipping \texttt{CFull} input length and are shared by both backbones. Gains range from 6.15\% to 7.95\% for LLaDA and from 7.23\% to 9.47\% for Llama. Among the 9,440 targets without recorded mixed-role commands, MC gains are 6.90\% and 8.31\%, respectively (Appendix~\ref{sec:appendix_context_sensitivity}). The benefits therefore extend across length groups and beyond recorded category-boundary cases. These groups are observational and can differ in target difficulty and dependency composition.

\subsection{Representation Responses to Context}
\label{sec:representation_analysis}

We complement these proof-level observations with the reference-conditioned representation probe defined in Section~\ref{sec:resource_comparisons}. Figure~\ref{fig:cka_heatmaps} reports block-32 dissimilarities across the four inference views. Adding MC to \texttt{CEnv} produces larger dissimilarity after \texttt{CFull} training than after \texttt{CEnv} training: 0.1038 versus 0.0377 for Llama and 0.1204 versus 0.0462 for LLaDA. Adding EC to \texttt{CMath} also produces larger representation responses after training that includes EC.

With each target's reference proof held fixed, checkpoints trained with a context category show larger representation changes when that category is supplied. The probe measures sensitivity under reference-proof conditioning. Verified-success and matched-replacement comparisons assess declaration benefits and selected-content advantages. Complete layer-wise results and model-specific readouts appear in Appendix~\ref{sec:appendix_cka}.


\subsection{Dependency Diagnostics and Selection}
\label{sec:dependency_selection_results}

The independent 11,064-target coverage evaluation examines which target dependencies reach final inputs (Table~\ref{tab:target_coverage_main}). Among targets with computable coverage, 58.26\% for LLaDA and 57.76\% for Llama lack complete coverage. Complete-coverage groups obtain larger conditional MC gains than zero-coverage groups. These observational groups can differ in difficulty and dependency composition; the omissions nevertheless identify a concrete opportunity to adjust selection. Complete coverage groups and input statistics are reported in Appendix~\ref{sec:appendix_final_input_coverage}, Table~\ref{tab:target_coverage_complete}. Dependency retention before tokenizer-specific clipping is reported separately in Appendix~\ref{sec:appendix_construction_retention}.

\begin{table}[H]
    \centering
    \small
    \setlength{\tabcolsep}{2.5pt}
    \setlength{\aboverulesep}{0pt}
    \setlength{\belowrulesep}{0pt}
    \setlength{\extrarowheight}{1pt}
    \begin{tabular}{lrrrr}
        \toprule
        \rowcolor{LCHeaderGray}
        & \multicolumn{2}{c}{
            LLaDA~{\color{LCVenueGray}\scriptsize NeurIPS'25}}
        & \multicolumn{2}{c}{
            Llama~{\color{LCVenueGray}\scriptsize arXiv'24}} \\
        \cmidrule(lr){2-3}
        \cmidrule(lr){4-5}
        \rowcolor{LCHeaderGray}
        Coverage & Targets & $\Delta_{M\mid E}$
        & Targets & $\Delta_{M\mid E}$ \\
        \midrule
        Zero     & 2,500 & +4.00 & 2,460 & +4.88 \\
        Partial  & 2,650 & +6.79 & 2,640 & +8.33 \\
        Complete & 3,690 & +7.72 & 3,730 & +9.12 \\
        \bottomrule
    \end{tabular}
    \caption{\textbf{Independent final-input coverage
    evaluation on 11,064 targets.}
    Coverage is computable for 8,840 LLaDA and
    8,830 Llama targets. Gains are Pass@32 differences
    in \%, computed before rounding.}
    \label{tab:target_coverage_main}
\end{table}

Context benefits also persist when selection excludes target reference proofs: a separate reselection control retains MC gains of 5.4\% and 6.4\% (Appendix~\ref{sec:reference_free_selection_results}). Because rebuilding context can change both MC and EC, the dependency-priority experiment uses a separate common baseline, shared candidate pool, and frozen EC to test the priority change under fixed prover conditions.

Prioritizing target dependencies raises coverage and improves Pass@32 by 1.40\% for LLaDA and 1.70\% for Llama, with mean inputs increasing by 4.4 and 4.9 tokens (Table~\ref{tab:dependency_priority_main}).

\begin{table}[H]
    \centering
    \small
    \setlength{\tabcolsep}{1.5pt}
    \setlength{\aboverulesep}{0pt}
    \setlength{\belowrulesep}{0pt}
    \setlength{\extrarowheight}{1pt}
    \begin{tabular}{llrrrr}
        \toprule
        \rowcolor{LCHeaderGray}
        Model & Policy & @1 & @32 & Cov. & Tokens \\
        \midrule
        LLaDA~{\color{LCVenueGray}\scriptsize NeurIPS'25}
            & Baseline
            & 5.94 & 43.90 & 47.70 & 696.8 \\
            & \textbf{Priority}
            & \textbf{6.17} & \textbf{45.30}
            & \textbf{72.26} & 701.2 \\
        \midrule
        Llama~{\color{LCVenueGray}\scriptsize arXiv'24}
            & Baseline
            & 6.25 & 44.55 & 48.88 & 714.2 \\
            & \textbf{Priority}
            & \textbf{6.53} & \textbf{46.25}
            & \textbf{73.48} & 719.1 \\
        \bottomrule
    \end{tabular}
    \caption{\textbf{Dependency-priority selection with a shared candidate pool and frozen EC.} Pass@1 and Pass@32 retain all 2,000 targets. Cov. is macro target coverage (\%) over 1,568 eligible LLaDA targets and 1,565 eligible Llama targets. Tokens is mean submitted-input length. Bold marks improvements over Baseline.}
    \label{tab:dependency_priority_main}
\end{table}

Coverage rises by 24.56\% and 24.60\% for LLaDA and Llama, respectively, with mean input growth below five tokens. These gains demonstrate the value of declaration selection within a fixed context budget.

Both policies operate without target reference proofs and retain identical checkpoints and inference budgets. The change gains 76 solved targets and loses 48 for LLaDA, and 88 and 54 for Llama (Table~\ref{tab:selector_paired_outcomes}). Additional reference-proof dependency coverage falls by 2.23\% and 1.70\% in the post-selection audit (Table~\ref{tab:selection_tradeoffs}).

The candidate-pool audit identifies a further limit on this selection change. Among 1,600 targets with complete, nonempty target-dependency mappings, the shared pool contains all dependencies for 1,410 targets, some for 80, and none for 110. Reranking cannot recover declarations absent from the pool. This audit distinguishes candidate-pool omissions from losses during selection and budget handling. These 1,600 targets form a separate audit population from the tokenizer-specific final-coverage groups. Further selection diagnostics are provided in Appendix~\ref{sec:appendix_dependency_priority}.

The selection change reallocates limited input space between target dependencies and other proof resources. For the evaluated checkpoints and 2,000-target subset, dependency priority improves overall proof success while changing which targets are solved. These results show how \textsc{LeanContext} can guide and test changes to declaration supply under fixed prover conditions.

The matched replacement results show that the advantage of selected declarations depends on prover training (Table~\ref{tab:matched_content_results}). Moving from \texttt{CEnv} to \texttt{CFull} training increases this advantage by 5.3\% for LLaDA and 6.3\% for Llama. This supports evaluating declaration selection together with the intended prover and its training condition.

Target-type coverage and verified success describe different aspects of context quality. The zero-coverage groups still obtain conditional MC gains of 4.00\% and 4.88\% for LLaDA and Llama, respectively (Table~\ref{tab:target_coverage_main}). The priority policy's gains and losses on individual targets further support reporting paired outcomes alongside aggregate success.

\section{Conclusion}
\label{sec:conclusion}

We introduced \textsc{LeanContext}, a benchmark of 129,826 Lean whole-proof generation problems that organizes complete mathematical declarations as MC and accompanying source commands as EC. Across representative diffusion-based and autoregressive-based methods, checkpoints trained with MC show larger conditional proof-generation gains from supplied declarations and larger advantages for selected declarations over matched replacements. Verified proofs also record reuse of selected MC on newly solved targets.
Final-input dependency audits guide a selection change that improves Pass@32 by 1.40\% and 1.70\% for LLaDA and Llama on the evaluated 2,000-target subset, without access to target reference proofs. The policy gains and loses solved targets and reduces coverage of additional reference-proof dependencies, showing the tradeoffs in allocating limited context. These findings support evaluating declaration selection together with prover training and testing improvements to context selection.


\section*{Limitations}
Our evaluations use a fixed Mathlib4 snapshot and 1,024-token input and output budgets. Extending these comparisons to other formal projects, longer contexts, and additional prover families would help characterize their broader applicability.

\textsc{LeanContext} compares complete source commands. Future work could identify compact target- and prover-specific MC and EC subsets or compressed representations that retain useful declaration content and local setting cues while preserving verified proof performance.

Adaptive selection policies could allocate the available budget between target dependencies, additional proof resources, and EC. Jointly adapting selection and prover training may help models use compact context more effectively. Analyses of generation trajectories could explain how declaration benefits and representation responses develop during proof construction.

\subsection*{Ethics Statement}
This work develops a benchmark for formal theorem proving from publicly available Mathlib source code and evaluates language models under controlled conditions. It does not involve human-subject experiments or the collection of new personal data. Proof candidates are checked by Lean in the original verification environment, and reported success requires passing this verification. The benchmark is intended to support reproducible research on the use of mathematical declarations in proof generation.



\bibliography{references}

\clearpage
\input{appendix}

\end{document}

%% file: appendix.tex
\onecolumn
\appendix
\raggedbottom
\newlength{\LCAppendixColumnWidth}
\setlength{\LCAppendixColumnWidth}{\dimexpr(\textwidth-\columnsep)/2\relax}
\setlength{\multicolsep}{0pt}
\setlength{\premulticols}{2\baselineskip}
\setlength{\postmulticols}{0pt}
\raggedcolumns
\newcommand{\LCAppendixSectionGap}{%
  \par\Needspace{5\baselineskip}\addvspace{12pt}}
\captionsetup{hypcap=false}
\setlength{\abovedisplayskip}{5pt plus 1pt minus 1pt}
\setlength{\belowdisplayskip}{5pt plus 1pt minus 1pt}
\setlength{\abovedisplayshortskip}{3pt plus 1pt minus 1pt}
\setlength{\belowdisplayshortskip}{3pt plus 1pt minus 1pt}

\makeatletter
\@ifpackageloaded{lineno}{%
  \renewcommand{\makeLineNumber}{%
    \logtheLineNumber\getLineNumber
    \docolaction{\makeLineNumberLeft}{\makeLineNumberLeft}{\makeLineNumberRight}}%
}{\providecommand{\linenumbers}{}\providecommand{\nolinenumbers}{}}
\makeatother

\newsavebox{\LCAppendixBox}
\newenvironment{LCAppendixBlock}{%
  \par\vspace{4pt}\begingroup\nolinenumbers
  \setlength{\abovecaptionskip}{3pt}
  \setlength{\belowcaptionskip}{1pt}
  \begin{lrbox}{\LCAppendixBox}\begin{minipage}{\textwidth}\centering
}{%
  \end{minipage}\end{lrbox}%
  \Needspace{\dimexpr\ht\LCAppendixBox+\dp\LCAppendixBox+2pt\relax}%
  \noindent\usebox{\LCAppendixBox}\par\endgroup\vspace{2pt}%
}
\nolinenumbers

\linenumbers
\begin{multicols}{2}
\section{Dataset Construction and Statistics}
\label{sec:appendix_construction}

This appendix documents source extraction, context selection, command classification, and the serialization of the four aligned views. It provides a verified selection example, an actual dataset-record excerpt, and the current selection prompt. Availability and command-count statistics describe the stored context before model-specific budget handling; experimental settings are specified in Appendix~\ref{sec:appendix_experiments}.

\subsection{Source Snapshot and Target Extraction}
\label{sec:appendix_source_extraction}

\textsc{LeanContext} contains 129,826 whole-proof generation problems from 4,469 Mathlib4~\citep{mathlib2020} source files. The construction protocol records Lean 4.14.0-rc1 and Mathlib4 commit \texttt{a7fc949f1b05c2a01e01}\allowbreak\texttt{c027fd9f480496a1253e}. Each problem corresponds to a theorem, lemma, or example command and retains its original source file, command identity, and source span.

The extractor records complete source commands together with their kinds, declaration names, source spans, and source order. It separates each target statement from its reference proof and draws context candidates from preceding commands in the same file. The target command and all subsequent commands are excluded. The public statement anonymizes the target name, while \texttt{original\_name} and the source metadata preserve its identity. Figure~\ref{fig:appendix_dataset_construction} summarizes the pipeline.

\subsection{Context Selection and Lean Verification}
\label{sec:appendix_original_selection}

Original context construction uses source structure, name-dependency evidence, and a language-model selection stage. The parser supplies candidate commands, and the selector returns identifiers from this pool. The current template prioritizes commands needed or strongly useful for parsing, name resolution, local variables, typeclasses, notation, attributes, or direct and indirect local dependencies. It excludes the target command and prohibits retaining commands solely because they appear earlier in the file.

Figure~\ref{fig:appendix_selection_record} shows the selection record for \texttt{fst\_list\_prod} in \texttt{Mathlib/\allowbreak{}Algebra/\allowbreak{}TrivSqZeroExt.\allowbreak{}lean}. The record lists 11 candidates and six kept identifiers, with model identifier \texttt{deepseek-v4-flash} and method \texttt{parser\_\allowbreak{}candidates\_\allowbreak{}plus\_\allowbreak{}llm\_\allowbreak{}minimized}. The example passed Lean verification, as recorded by \texttt{verified\_\allowbreak{}by\_\allowbreak{}lean: true}. The prompt uses \emph{minimal} as a selection objective.

Candidate construction and original selection can inspect the target reference proof. In the reconstructed request, \texttt{target.text} contains both the statement and proof, and the stored selection reason mentions \texttt{fst\_one} and \texttt{fst\_mul}, which the proof uses. We therefore label this construction regime \texttt{original\_\allowbreak{}reference\_\allowbreak{}assisted}. The selection controls that exclude reference-proof access are specified in Appendix~\ref{sec:appendix_dependencies}.

Selection output is followed by filtering and restoration of source order. The recorded post-selection rule retains directly parsed theorem and lemma commands only when their declaration names occur lexically in the target statement. The six identifiers in the selection record and the four commands in the final dataset record describe different stages. The final record retains \texttt{fst} and three EC commands; \texttt{fst\_one} and \texttt{fst\_mul} are absent from its stored context.

\subsection{Command Classification}
\label{sec:appendix_command_classification}

Retained commands are assigned to MC or EC by effective command kind, as summarized in Table~\ref{tab:appendix_command_partition}. For supported wrapper forms, the extractor recognizes the enclosed declaration kind; otherwise, it retains the parser kind. MC preserves complete mathematical declarations, including their signatures and existing bodies. EC contains the remaining retained source commands. An instance declaration belongs to MC, including a local instance, whereas a \texttt{variable} command containing typeclass binders remains EC.

The final \texttt{fst\_list\_prod} record contains one MC declaration, \texttt{fst} (\texttt{cmd\_000012}), and three EC commands: \texttt{universe} (\texttt{cmd\_000002}), local notation (\texttt{cmd\_000004}), and \texttt{namespace} (\texttt{cmd\_000006}). The \texttt{mathematical\_context} and \texttt{environment\_context} arrays store these groups separately, while \texttt{full\_context} preserves their combined source order.

\subsection{Dataset Records and Aligned Views}
\label{sec:appendix_dataset_records}

Figure~\ref{fig:appendix_dataset_record} presents an excerpt of test sample \texttt{cc1c0999b59e1d4f78e277ee}. The \texttt{statement} field stores the anonymized target statement ending in \texttt{:=}, and \texttt{target} stores the reference proof. The record also contains \texttt{split}, \texttt{proof\_family}, target identity, source metadata, and \texttt{verification\_\allowbreak{}location}. Here, \texttt{proof\_family} is \texttt{Strict Term}. The \texttt{source\_revision} object records the Mathlib release, Lean toolchain, recorded commit, the basis of that commit record, and the source-file SHA-256 hash.

Each context item retains its \texttt{text}, \texttt{command\_id}, \texttt{command\_uid}, \texttt{source\_span}, \texttt{parser\_kind}, and \texttt{effective\_kind}. The UID combines the source-file path with the command identifier. The \texttt{position} field is an ordinal in the combined retained context, whereas \texttt{source\_order} records the original command order. For example, \texttt{fst} retains \texttt{position: 3} and \texttt{source\_order: 11}, even when it is the only command in \texttt{CMath}. Source spans retain the original byte offsets and line and column coordinates.

The \texttt{views} object contains \texttt{C0}, \texttt{CMath}, \texttt{CEnv}, and \texttt{CFull}. Each view shares the sample identity, statement, and reference proof and records its own \texttt{prompt}, \texttt{context\_items}, \texttt{context\_ids}, and \texttt{context\_uids}. Context items additionally record their \texttt{role} as MC or EC. Table~\ref{tab:appendix_actual_views} gives the actual command sequences for this example.

The stored \texttt{prompt} concatenates visible command texts in source order, separates them with blank lines, and appends the shared statement. Thus, \texttt{C0} contains only the statement, \texttt{CMath} adds the \texttt{fst} definition, \texttt{CEnv} adds the three EC commands, and \texttt{CFull} includes all four. These raw prompt strings do not contain MC/EC section headings. Model-specific input formatting and budget handling are described in Appendix~\ref{sec:appendix_experiments}.

For training, \texttt{prompt} supplies the input and \texttt{target} supplies the reference-proof answer. At evaluation, the reference proof is retained for annotation and analysis and is excluded from the generation input. Generated proofs are checked at the recorded original location with the complete Lean environment, regardless of which view is shown to the model.

\subsection{Selection Prompt and Reconstructed Request}
\label{sec:appendix_selection_prompt}

Figure~\ref{fig:appendix_selection_prompt} presents the current system prompt and a reconstructed user request. The request contains \texttt{task}, \texttt{source\_file}, \texttt{source\_module}, \texttt{target}, and \texttt{candidate\_commands}. Each candidate records its identifier, kind, name, line range, and complete text. The name may be \texttt{null}, as in an unnamed instance or notation command. Here, \texttt{target.text} is the complete source command used for selection; it differs from the reference-proof \texttt{target} field in the final dataset record.

The prompt requests \texttt{keep\_command\_ids}, \texttt{drop\_command\_ids}, and \texttt{reason}, whereas the stored selection metadata uses \texttt{kept\_command\_ids} and \texttt{selection\_reason}. The user request is reconstructed from the current template, the stored original candidate identifiers, and the original parser command text. It documents the request format and is not an archived historical API request.

\subsection{Dataset Splits}
\label{sec:appendix_dataset_splits}

The fixed split uses seed 20260811. Before splitting, target names are anonymized and consecutive whitespace in public target statements is collapsed. Source files connected by a repeated normalized statement are grouped and assigned to the same partition, preserving both source-file and normalized-statement disjointness.

Normalization handles target names and whitespace; it does not equate arbitrary binder renamings or mathematically equivalent statements. All experiments retain the original target identifiers and split assignments. The matched-replacement and selection-control subsets are specified in Appendices~\ref{sec:appendix_matched} and~\ref{sec:appendix_dependencies}, respectively, within these fixed partitions.

\subsection{Dataset Statistics and Source Audit}
\label{sec:appendix_dataset_statistics}

Table~\ref{tab:context_availability_complete} reports context availability before model-specific budget handling. A category is nonempty when it contains at least one selected command. The Both category therefore describes the saved context before clipping, rather than complete retention in the final model input. The main context comparisons use its 11,064 test targets.

Table~\ref{tab:appendix_selected_command_counts} gives selected-command counts for Both records. Each occurrence is a target--command pair, so a source command selected for different targets contributes once per target. Means describe the number of selected commands per target.

Command counts and token lengths are measured separately. Under the LLaDA tokenizer on the 11,064 Both test targets, selected MC text has a pre-clipping mean and median of 450 and 200 tokens; selected EC text has a mean and median of 140 and 90 tokens. These lengths describe context text only. Submitted inputs also include the target statement and model-specific special tokens and are checked against the 1,024-token input limit with the actual model tokenizer.

The lineage audit covers all 129,826 targets across 4,469 source files. All 766,839 selected-command occurrences precede their targets, exclude the target command, and match their parser text. A separate comparison against the pinned Mathlib source checks 238,839 source commands and all 73,755 reference-proof bodies in the Both population against their original source spans.

These counts use different units: 766,839 counts selected-command occurrences across the full dataset; 238,839 counts source commands checked in the separate source comparison; and 73,755 counts reference-proof records in the Both population. The source audit establishes source identity and ordering, while Lean verification reports the validity of the checked construction.
\end{multicols}
\nolinenumbers

\begin{LCAppendixBlock}
\centering
    \includegraphics[width=\textwidth]{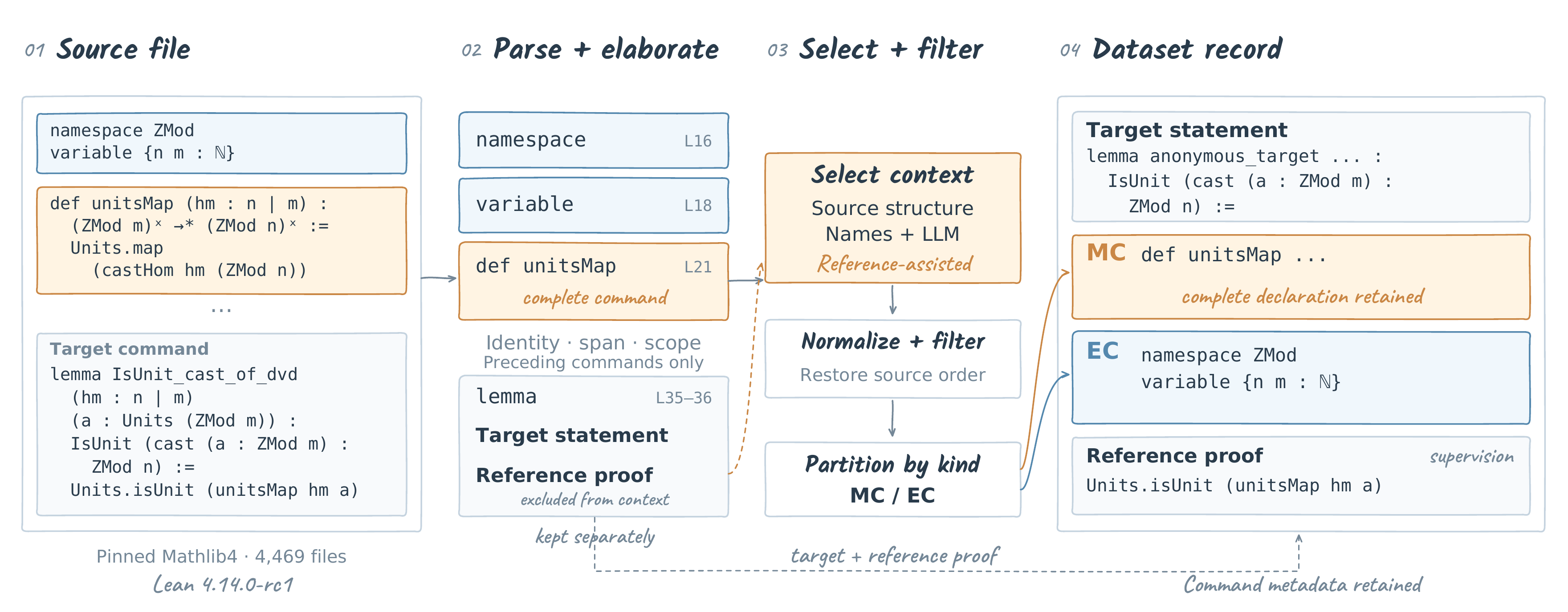}
    \captionof{figure}{\textbf{Dataset construction from Lean source commands.} The pipeline extracts target and preceding-command records, selects and filters context, and organizes retained commands into MC and EC while preserving source identities and spans.}
    \label{fig:appendix_dataset_construction}
\end{LCAppendixBlock}

\begin{LCAppendixBlock}
\centering
    \includegraphics[width=\textwidth]{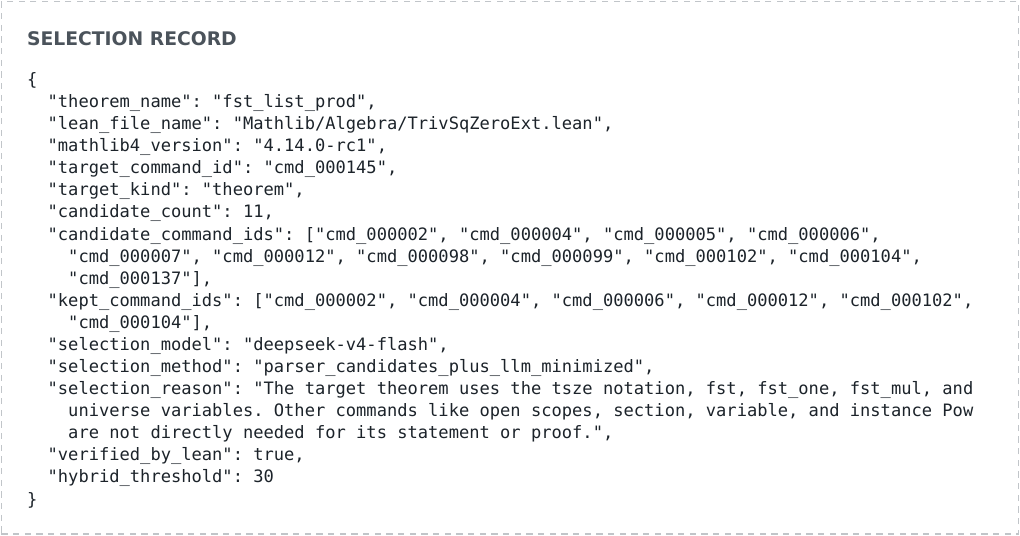}
    \captionof{figure}{\textbf{Context-selection metadata for \texttt{fst\_list\_prod}, with \texttt{verified\_by\_lean: true}.} Kept identifiers describe the selection output; the final stored context is shown in Figure~\ref{fig:appendix_dataset_record}.}
    \label{fig:appendix_selection_record}
\end{LCAppendixBlock}

\begin{LCAppendixBlock}
\centering
    \includegraphics[width=0.94\textwidth,height=0.72\textheight,keepaspectratio]{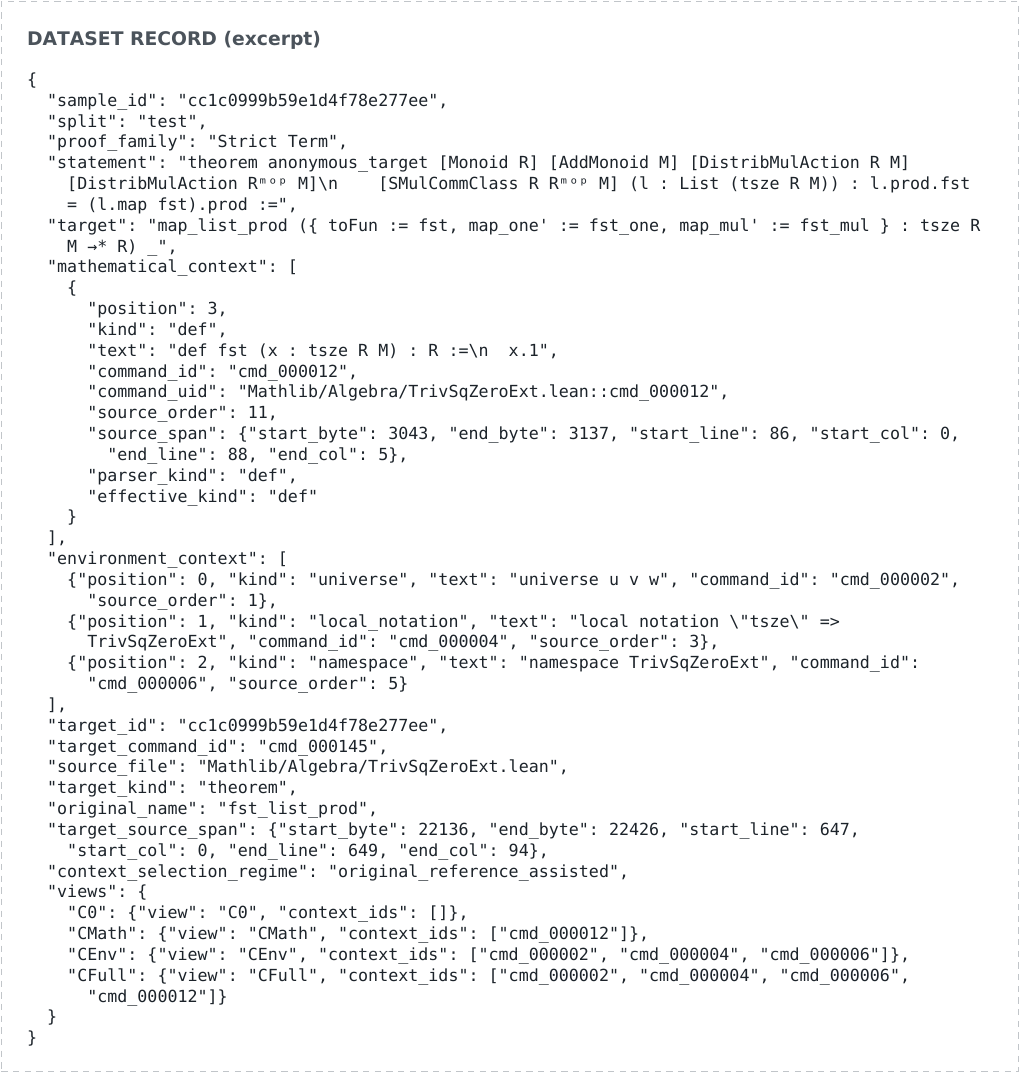}
    \captionof{figure}{\textbf{An excerpt of the actual dataset record for \texttt{fst\_list\_prod}.} The shared statement and reference proof are shown once. The excerpt preserves the MC command metadata, the three EC commands, and the context identifiers of all four views; repeated metadata and full prompt strings are omitted for readability.}
    \label{fig:appendix_dataset_record}
\end{LCAppendixBlock}

\begin{LCAppendixBlock}
\noindent
\begin{minipage}[t]{\LCAppendixColumnWidth}
\vspace{0pt}
\centering
    \small
    \setlength{\tabcolsep}{3pt}
    \setlength{\aboverulesep}{0pt}
    \setlength{\belowrulesep}{0pt}
    \setlength{\extrarowheight}{1pt}
    \renewcommand{\arraystretch}{1.0}

    \begin{tabular}{@{}ll@{}}
        \toprule
        \rowcolor{LCHeaderGray}
        \textbf{Category} & \textbf{Effective kinds or representative commands} \\
        \midrule
        MC & \texttt{theorem}, \texttt{lemma} \\
           & \texttt{def}, \texttt{abbrev}, \texttt{opaque} \\
           & \texttt{axiom}, \texttt{constant} \\
           & \texttt{structure}, \texttt{class}, \texttt{inductive} \\
           & \texttt{instance}, \texttt{local\_instance}, \texttt{alias} \\
        \midrule
        EC & \shortstack[l]{All remaining retained commands, including \\
             \texttt{variable}, \texttt{universe}, \texttt{namespace}, \texttt{section}, \\
             \texttt{open}, notation, attributes, options, and \texttt{export}.} \\
        \bottomrule
    \end{tabular}
    \captionof{table}{\textbf{Command classification.} EC examples are illustrative. Supported wrappers follow the recognized declaration kind.}
    \label{tab:appendix_command_partition}
\end{minipage}\hfill
\begin{minipage}[t]{\LCAppendixColumnWidth}
\vspace{0pt}
\centering
    \small
    \setlength{\tabcolsep}{3pt}
    \setlength{\aboverulesep}{0pt}
    \setlength{\belowrulesep}{0pt}
    \setlength{\extrarowheight}{1pt}
    \renewcommand{\arraystretch}{1.0}

    \begin{tabular}{@{}lll@{}}
        \toprule
        \rowcolor{LCHeaderGray}
        \textbf{View} & \textbf{Context} & \textbf{Command identifiers} \\
        \midrule
        \texttt{C0} & None & \texttt{[]} \\
        \texttt{CMath} & MC & \texttt{cmd\_000012} \\
        \texttt{CEnv} & EC & \shortstack[l]{
        \texttt{cmd\_000002}, \texttt{cmd\_000004}, \\
        \texttt{cmd\_000006}} \\
        \texttt{CFull} & MC and EC & \shortstack[l]{
        \texttt{cmd\_000002}, \texttt{cmd\_000004}, \\
        \texttt{cmd\_000006}, \texttt{cmd\_000012}} \\
        \bottomrule
    \end{tabular}
    \captionof{table}{\textbf{Actual context sequences in the four stored views of \texttt{fst\_list\_prod}.} All views share the same target statement and reference proof.}
    \label{tab:appendix_actual_views}
\end{minipage}
\end{LCAppendixBlock}

\begin{LCAppendixBlock}
\centering
    \includegraphics[width=0.94\textwidth,height=0.76\textheight,keepaspectratio]{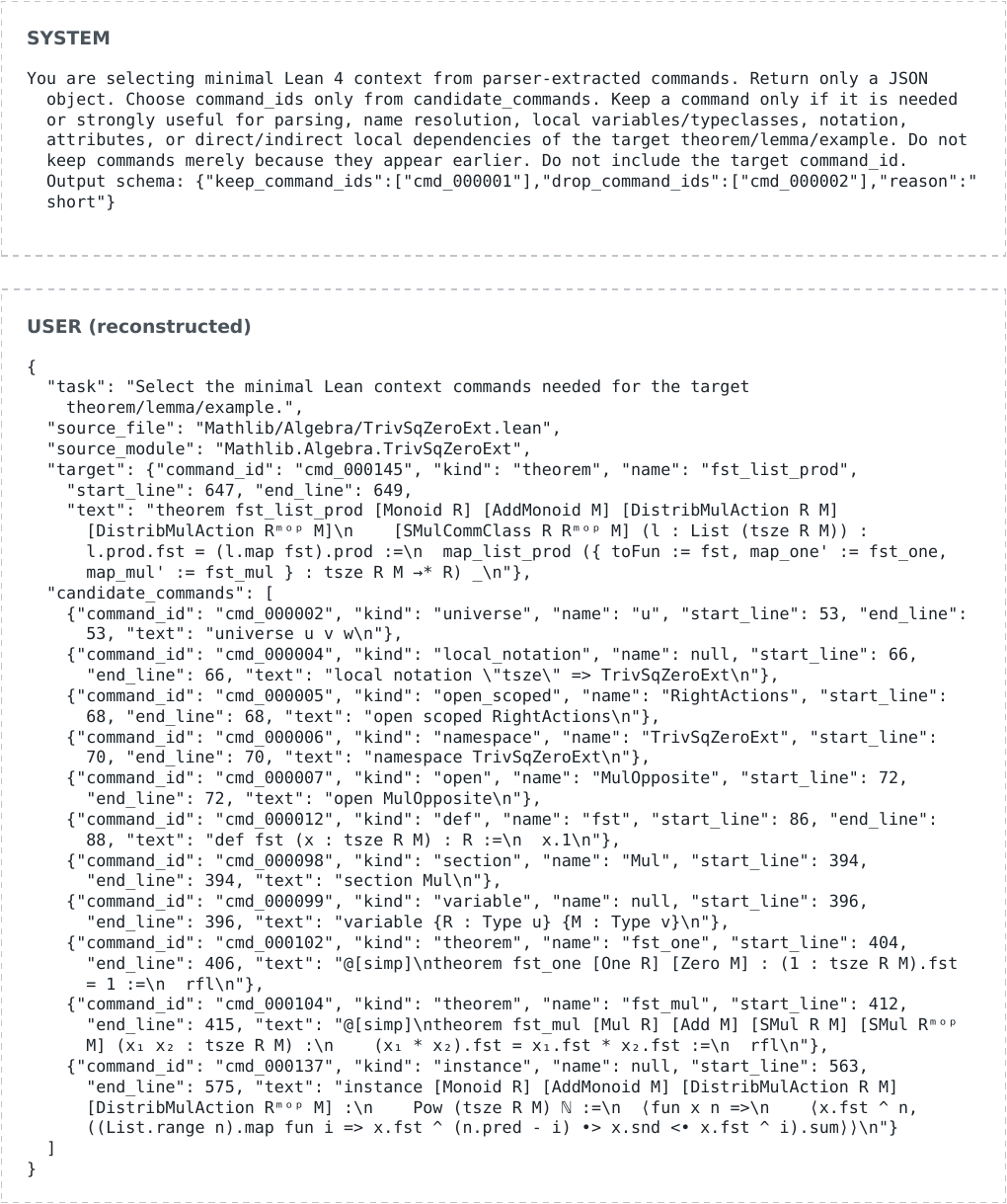}
    \captionof{figure}{\textbf{The current selection system prompt and the complete reconstructed user request for \texttt{fst\_list\_prod}.} The request uses the original candidate identifiers and parser command text.}
    \label{fig:appendix_selection_prompt}
\end{LCAppendixBlock}

\begin{LCAppendixBlock}
\noindent
\begin{minipage}[t]{\LCAppendixColumnWidth}
\vspace{0pt}
\centering
    \small
    \setlength{\tabcolsep}{3pt}
    \setlength{\aboverulesep}{0pt}
    \setlength{\belowrulesep}{0pt}
    \setlength{\extrarowheight}{1pt}
    \renewcommand{\arraystretch}{1.0}

    \begin{tabular}{@{}lrrrrr@{}}
        \toprule
        \rowcolor{LCHeaderGray}
        \textbf{Split} & \textbf{Targets} & \textbf{MC only} & \textbf{EC only} & \textbf{Neither} & \textbf{Both} \\
        \midrule
        Train & 90,879 & 401 & 38,248 & 602 & 51,628 \\
        Valid. & 19,473 & 86 & 8,195 & 129 & 11,063 \\
        Test & 19,474 & 85 & 8,196 & 129 & 11,064 \\
        \midrule
        Total & 129,826 & 572 & 54,639 & 860 & 73,755 \\
        \bottomrule
    \end{tabular}
    \captionof{table}{\textbf{Dataset splits and selected-context availability before input clipping.}}
    \label{tab:context_availability_complete}
\end{minipage}\hfill
\begin{minipage}[t]{\LCAppendixColumnWidth}
\vspace{0pt}
\centering
    \small
    \setlength{\tabcolsep}{3pt}
    \setlength{\aboverulesep}{0pt}
    \setlength{\belowrulesep}{0pt}
    \setlength{\extrarowheight}{1pt}
    \renewcommand{\arraystretch}{1.0}

    \begin{tabular}{@{}lrrrrr@{}}
        \toprule
        \rowcolor{LCHeaderGray}
        \textbf{Split} & \textbf{Both} & \textbf{MC total} & \textbf{MC mean} & \textbf{EC total} & \textbf{EC mean} \\
        \midrule
        Train & 51,628 & 109,886 & 2.13 & 255,876 & 4.96 \\
        Valid. & 11,063 & 23,057 & 2.08 & 55,407 & 5.01 \\
        Test & 11,064 & 23,422 & 2.12 & 57,594 & 5.21 \\
        \midrule
        Total & 73,755 & 156,365 & 2.12 & 368,877 & 5.00 \\
        \bottomrule
    \end{tabular}
    \captionof{table}{\textbf{Selected-command occurrences on Both records before input clipping.} Totals count occurrences across records; means are per target.}
    \label{tab:appendix_selected_command_counts}
\end{minipage}
\end{LCAppendixBlock}

\LCAppendixSectionGap
\linenumbers
\begin{multicols}{2}
\section{Experimental Settings and Additional Results}
\label{sec:appendix_experiments}
\label{sec:appendix_results}

This appendix documents input construction and candidate handling for the controlled LLaDA and Llama runs, then reports additional evaluations across model variants, sampling budgets, temperatures, and test populations.

\subsection{Input Construction and Proof Generation}
\label{sec:appendix_input_pipeline}

The input and per-attempt output budgets are each 1,024 tokens. Target statements and reference proofs are retained without truncation. For retained samples, when a serialized input exceeds the input budget, we remove complete context commands from the end of the source-ordered context until the input fits. Commands are removed as whole units, preserving the complete statement and the text of every retained command. The four views use the same budget-handling rules and a common sample filter to preserve target alignment across views.

We arrange selected context commands in their original source order, concatenate them with the target statement, and retain a final newline. The prompt contains no additional instructions, chat templates, or section headings. Tokenization uses fixed versions of the LLaDA and Llama-3.1 tokenizers. Llama inputs begin with exactly one BOS token. For supervised training, the reference proof follows the prompt and ends with exactly one EOS token. In proof-generation evaluations, the model receives the context--statement prompt.

Autoregressive generation stops when EOS is produced or the output-token limit is reached. For LLaDA, extraction takes place after denoising completes and retains the content before the first EOS. The inference and verification implementation includes functions for code-block cleanup, proof-body extraction, and locating and replacing the original Lean declaration. These functions prepare the generated proof body for verification at the original target location.

\subsection{Success Metrics}
\label{sec:appendix_success_metrics}

For $N$ evaluation targets and $c_i$ verified candidates among 32 attempts for target $i$, we estimate sampled Pass@$k$ as~\citep{chen2021evaluating}
\begin{equation}
    \widehat{\mathrm{Pass@}k}
    = \frac{1}{N}\sum_{i=1}^{N}
      \left(1-
      \frac{\binom{32-c_i}{k}}{\binom{32}{k}}\right),
    \label{eq:sampled_passk}
\end{equation}
for $k\in\{1,8,16,32\}$. The numerator is zero when $32-c_i<k$. Targets without submitted inputs have $c_i=0$ and remain in the denominator. Sampled Pass@1 averages verified candidate fractions, whereas Pass@32 measures the fraction of targets with at least one verified candidate. Greedy accuracy, denoted G, uses a separate deterministic attempt. Scores and absolute score differences are reported in \%. All reported scores are percentages.


\raggedbottom

\subsection{Model Variants and Adaptation Settings}
\label{sec:appendix_model_variants}

Tables~\ref{tab:fixed_context_0_math} and~\ref{tab:fixed_context_env_full} retain the complete fixed-checkpoint results for all four inference views. In these fixed-checkpoint tables, SFT and LoRA~\citep{hu2022lora} rows use \texttt{CFull} training. The additional \texttt{C0} SFT rows use checkpoints trained without local context. Each checkpoint remains fixed across its four inference conditions.

The adapted families include LLaDA and Llama~\citep{nie2025large,llama2024herd}, LLaDA-1.5~\citep{zhu2026llada15}, Dream~\citep{ye2025dream}, and Qwen variants~\citep{qwen2025qwen3,qwen2024qwen25,yang2024qwen25math,hui2024qwen25coder}. Released Lean provers include DeepSeek-Prover~\citep{xin2025deepseekprover15,ren2025deepseekprover2}, Goedel-Prover~\citep{lin2025goedelprover}, and TheoremLlama~\citep{wang2024theoremllama}. Released general-purpose models and GPT-4o/GPT-4.1 API results provide additional reference conditions.

For the Base models adapted with LoRA, adding MC to EC increases Pass@32 by 5.4\% for LLaDA and 5.6\% for Llama. The corresponding gains are 7.1\% and 8.5\% after SFT. All five released Lean provers also obtain positive conditional MC gains, ranging from 1.6\% to 5.1\%. These evaluations extend the observed context benefits across the recorded settings; differences in training histories prevent interpreting absolute model rankings as isolated architecture or adaptation effects.

\subsection{Sampling-Budget Effects}
\label{sec:appendix_sampling_budget}

The complete tables report Pass@1, Pass@8, Pass@16, and Pass@32 from the same 32 candidate outcomes, allowing comparisons across attempt budgets without resampling. For the \texttt{CFull}-trained primary checkpoints, conditional MC gains remain positive at every reported budget. They increase from 1.0\% to 7.1\% for LLaDA and from 1.3\% to 8.5\% for Llama between Pass@1 and Pass@32.

Relative rankings also depend on the budget. Under \texttt{CFull} inference, DeepSeek-Prover-V2-7B obtains Pass@8 of 33.0\%, compared with 27.1\% and 27.9\% for the primary LLaDA and Llama checkpoints. At Pass@16, the primary checkpoints obtain 37.8\% and 38.6\%, compared with 36.4\% for DeepSeek-Prover-V2-7B. These comparisons describe the evaluated checkpoints under matched attempt counts.

\subsection{Temperature Sensitivity}
\label{sec:appendix_temperature}

Tables~\ref{tab:temperature_0_math} and~\ref{tab:temperature_env_full} report the completed temperature experiments on the same 11,064 targets. The primary LLaDA and Llama checkpoints use SFT with \texttt{CFull} training; DeepSeek-Prover-V1.5-RL and Goedel-Prover-SFT are released checkpoints. Temperatures 0.2, 0.6, and 1.0 each use 32 attempts. Temperature 0 reports a separate greedy attempt.

\texttt{CFull} obtains the highest Pass@32 across inference views in every tested sampling configuration, and the highest greedy score. The partial-view ordering varies: for Llama, \texttt{CMath} and \texttt{CEnv} tie at temperature 0.2, \texttt{CMath} leads at 0.6, and \texttt{CEnv} leads at 1.0. The combined-context benefit therefore persists across the tested temperatures, while category comparisons remain specific to the decoding condition.

\subsection{Matched Training and Evaluation}
\label{sec:appendix_matched_view_results}

Tables~\ref{tab:matched_training_C0}--\ref{tab:matched_training_CFull} reproduce the complete matched training and evaluation results. Each input view is evaluated using a checkpoint trained with that view. Comparisons between the tables therefore change both training and inference context; they are not fixed-checkpoint ablations. Results on the complete test split and the both-category subset are reported as separate column groups. Each table retains greedy accuracy and Pass@1, Pass@8, Pass@16, and Pass@32.

On the full test split, matched \texttt{CFull} training and evaluation obtain Pass@32 of 50.8\% for LLaDA and 52.0\% for Llama under SFT. Across the 16 adaptation configurations, the \texttt{CFull} scores exceed matched \texttt{C0} scores by 6.2\%--9.2\%. These scores describe training and evaluation with the same view, separately from the fixed-checkpoint comparisons.
\end{multicols}
\nolinenumbers

\begin{LCAppendixBlock}
\centering
    \small
    \setlength{\tabcolsep}{4pt}
    \setlength{\aboverulesep}{0pt}
    \setlength{\belowrulesep}{0pt}
    \setlength{\extrarowheight}{1pt}
    \renewcommand{\arraystretch}{1.0}

    \begin{tabular}{@{}llrrrrrrrrrr@{}}
        \toprule
        \rowcolor{LCHeaderGray}
         &  & \multicolumn{5}{c}{\textbf{\texttt{C0}}} & \multicolumn{5}{c}{\textbf{\texttt{CMath}}} \\
        \cmidrule(lr){3-7} \cmidrule(lr){8-12}
        \rowcolor{LCHeaderGray}
        \textbf{Model} & \textbf{Condition} & \textbf{G} & \textbf{@1} & \textbf{@8} & \textbf{@16} & \textbf{@32} & \textbf{G} & \textbf{@1} & \textbf{@8} & \textbf{@16} & \textbf{@32} \\
        \midrule
        LLaDA-8B-Base & SFT & 5.6 & 4.4 & 17.4 & 24.2 & 31.2 & 7.2 & 5.7 & 22.3 & 31.1 & 40.1 \\
        Llama-3.1-8B & SFT & 6.1 & 4.7 & 18.4 & 25.5 & 32.8 & 7.8 & 6.0 & 23.5 & 32.6 & 41.8 \\
        LLaDA-8B-Base & LoRA & 5.0 & 3.9 & 15.9 & 22.6 & 29.8 & 5.9 & 4.7 & 19.5 & 27.6 & 36.4 \\
        Llama-3.1-8B & LoRA & 5.4 & 4.2 & 17.2 & 24.4 & 31.7 & 6.6 & 5.1 & 20.8 & 29.4 & 38.2 \\
        LLaDA-8B-Instruct & SFT & 6.5 & 5.1 & 19.8 & 27.2 & 34.6 & 8.1 & 6.4 & 24.9 & 34.2 & 43.5 \\
        Llama-3.1-8B-Instruct & SFT & 7.0 & 5.4 & 21.1 & 28.6 & 36.3 & 8.8 & 6.8 & 26.4 & 35.7 & 45.4 \\
        LLaDA-8B-Instruct & LoRA & 5.6 & 4.4 & 17.7 & 24.8 & 32.2 & 6.7 & 5.3 & 21.4 & 30.0 & 38.9 \\
        Llama-3.1-8B-Instruct & LoRA & 6.1 & 4.7 & 19.1 & 26.5 & 34.0 & 7.4 & 5.7 & 22.9 & 31.8 & 40.8 \\
        LLaDA-1.5 & LoRA & 8.4 & 6.6 & 25.2 & 33.4 & 42.1 & 9.7 & 7.7 & 29.6 & 39.1 & 49.3 \\
        Dream-v0-Base-7B & LoRA & 5.1 & 4.0 & 16.6 & 23.4 & 30.8 & 6.1 & 4.8 & 19.8 & 27.9 & 36.7 \\
        Dream-v0-Instruct-7B & LoRA & 6.0 & 4.7 & 18.5 & 25.8 & 33.4 & 7.1 & 5.6 & 22.3 & 31.1 & 40.2 \\
        Qwen3-8B-Base & LoRA & 7.0 & 5.4 & 21.2 & 29.0 & 37.1 & 8.2 & 6.3 & 25.0 & 34.1 & 43.6 \\
        Qwen3-8B & LoRA & 8.6 & 6.7 & 25.5 & 34.0 & 42.9 & 10.0 & 7.7 & 29.4 & 39.2 & 49.4 \\
        Qwen2.5-7B & LoRA & 6.6 & 5.1 & 20.0 & 27.3 & 34.9 & 7.8 & 6.0 & 23.6 & 32.2 & 41.2 \\
        Qwen2.5-Math-7B & LoRA & 7.1 & 5.5 & 21.5 & 29.4 & 37.6 & 8.6 & 6.6 & 26.2 & 35.8 & 45.8 \\
        Qwen2.5-Coder-7B & LoRA & 7.1 & 5.5 & 21.7 & 29.7 & 37.9 & 7.8 & 6.0 & 23.7 & 32.4 & 41.4 \\
        \midrule
        LLaDA-8B-Base & \texttt{C0} SFT & 6.1 & 4.8 & 20.4 & 29.4 & 38.6 & 6.2 & 4.9 & 20.7 & 29.8 & 39.1 \\
        Llama-3.1-8B & \texttt{C0} SFT & 6.7 & 5.2 & 21.4 & 30.5 & 39.8 & 6.9 & 5.3 & 21.8 & 31.1 & 40.6 \\
        \midrule
        DeepSeek-Prover-V1.5-SFT & Released & 10.6 & 11.5 & 17.6 & 19.4 & 20.7 & 13.6 & 14.7 & 22.4 & 24.7 & 26.4 \\
        DeepSeek-Prover-V1.5-RL & Released & 11.8 & 12.8 & 19.4 & 21.5 & 22.9 & 14.5 & 15.7 & 23.9 & 26.3 & 28.1 \\
        DeepSeek-Prover-V2-7B & Released & 15.2 & 16.4 & 25.0 & 27.5 & 29.4 & 18.3 & 19.8 & 30.2 & 33.4 & 35.6 \\
        Goedel-Prover-SFT & Released & 12.2 & 13.1 & 20.0 & 22.1 & 23.6 & 15.0 & 16.2 & 24.7 & 27.3 & 29.1 \\
        TheoremLlama & Released & 6.0 & 6.5 & 9.9 & 11.0 & 11.7 & 7.3 & 7.9 & 12.1 & 13.3 & 14.2 \\
        \midrule
        Llama-3.1-8B-Instruct & Released & 3.3 & 3.3 & 6.2 & 7.0 & 7.6 & 3.5 & 3.5 & 6.6 & 7.4 & 8.1 \\
        Qwen2.5-Math-7B-Instruct & Released & 5.5 & 5.5 & 10.4 & 11.7 & 12.8 & 7.6 & 7.6 & 14.3 & 16.1 & 17.6 \\
        Qwen2.5-Coder-7B-Instruct & Released & 6.5 & 6.5 & 12.1 & 13.6 & 14.9 & 7.2 & 7.2 & 13.6 & 15.3 & 16.7 \\
        \midrule
        GPT-4o & API & 8.0 & 8.0 & 16.0 & 18.3 & 19.8 & 10.5 & 10.5 & 21.1 & 24.1 & 26.1 \\
        GPT-4.1 & API & 8.7 & 8.7 & 17.4 & 19.9 & 21.6 & 11.4 & 11.4 & 22.9 & 26.1 & 28.3 \\
        \bottomrule
    \end{tabular}
    \captionof{table}{\textbf{Complete fixed-checkpoint results under \texttt{C0} and \texttt{CMath} on 11,064 targets.} G is a separate greedy attempt; @1, @8, @16, and @32 are sampled Pass@$k$. SFT and LoRA conditions use \texttt{CFull} training; \texttt{C0} SFT uses \texttt{C0} training. Released and API conditions receive no additional \textsc{LeanContext} training. DeepSeek-Prover-V2-7B uses non-CoT generation.}
    \label{tab:fixed_context_0_math}
\end{LCAppendixBlock}

\begin{LCAppendixBlock}
\centering
    \small
    \setlength{\tabcolsep}{4pt}
    \setlength{\aboverulesep}{0pt}
    \setlength{\belowrulesep}{0pt}
    \setlength{\extrarowheight}{1pt}
    \renewcommand{\arraystretch}{1.0}

    \begin{tabular}{@{}llrrrrrrrrrr@{}}
        \toprule
        \rowcolor{LCHeaderGray}
         &  & \multicolumn{5}{c}{\textbf{\texttt{CEnv}}} & \multicolumn{5}{c}{\textbf{\texttt{CFull}}} \\
        \cmidrule(lr){3-7} \cmidrule(lr){8-12}
        \rowcolor{LCHeaderGray}
        \textbf{Model} & \textbf{Condition} & \textbf{G} & \textbf{@1} & \textbf{@8} & \textbf{@16} & \textbf{@32} & \textbf{G} & \textbf{@1} & \textbf{@8} & \textbf{@16} & \textbf{@32} \\
        \midrule
        LLaDA-8B-Base & SFT & 7.7 & 5.9 & 23.2 & 32.3 & 41.6 & 9.0 & 6.9 & 27.1 & 37.8 & 48.7 \\
        Llama-3.1-8B & SFT & 7.9 & 5.9 & 23.1 & 32.0 & 41.1 & 9.5 & 7.2 & 27.9 & 38.6 & 49.6 \\
        LLaDA-8B-Base & LoRA & 6.5 & 5.0 & 20.6 & 29.2 & 38.5 & 7.4 & 5.7 & 23.5 & 33.3 & 43.9 \\
        Llama-3.1-8B & LoRA & 7.1 & 5.3 & 21.6 & 30.6 & 39.8 & 8.0 & 6.1 & 24.7 & 35.0 & 45.4 \\
        LLaDA-8B-Instruct & SFT & 8.6 & 6.6 & 25.7 & 35.2 & 44.8 & 9.9 & 7.6 & 29.3 & 40.2 & 51.2 \\
        Llama-3.1-8B-Instruct & SFT & 9.0 & 6.8 & 26.3 & 35.5 & 45.1 & 10.4 & 7.9 & 30.6 & 41.3 & 52.5 \\
        LLaDA-8B-Instruct & LoRA & 7.3 & 5.6 & 22.9 & 32.0 & 41.5 & 8.3 & 6.4 & 26.0 & 36.4 & 47.2 \\
        Llama-3.1-8B-Instruct & LoRA & 7.9 & 5.9 & 23.9 & 33.3 & 42.7 & 8.9 & 6.8 & 27.3 & 37.9 & 48.6 \\
        LLaDA-1.5 & LoRA & 10.6 & 8.1 & 30.8 & 40.8 & 51.4 & 11.6 & 8.9 & 34.0 & 45.0 & 56.7 \\
        Dream-v0-Base-7B & LoRA & 6.7 & 5.1 & 21.1 & 29.7 & 39.1 & 7.6 & 5.8 & 23.9 & 33.7 & 44.3 \\
        Dream-v0-Instruct-7B & LoRA & 7.7 & 5.9 & 23.3 & 32.5 & 42.1 & 8.7 & 6.7 & 26.5 & 37.0 & 47.9 \\
        Qwen3-8B-Base & LoRA & 8.9 & 6.7 & 26.3 & 35.9 & 45.9 & 9.7 & 7.4 & 29.4 & 40.1 & 51.3 \\
        Qwen3-8B & LoRA & 10.5 & 7.9 & 30.4 & 40.5 & 51.1 & 11.6 & 8.8 & 33.6 & 44.8 & 56.4 \\
        Qwen2.5-7B & LoRA & 8.4 & 6.3 & 25.0 & 34.2 & 43.7 & 9.3 & 7.1 & 28.0 & 38.3 & 48.9 \\
        Qwen2.5-Math-7B & LoRA & 8.0 & 6.0 & 23.6 & 32.3 & 41.3 & 9.2 & 7.0 & 27.8 & 38.0 & 48.5 \\
        Qwen2.5-Coder-7B & LoRA & 8.9 & 6.7 & 26.4 & 36.2 & 46.2 & 9.5 & 7.2 & 28.4 & 38.8 & 49.6 \\
        \midrule
        LLaDA-8B-Base & \texttt{C0} SFT & 6.1 & 4.7 & 20.0 & 28.8 & 37.8 & 6.3 & 4.8 & 20.3 & 29.1 & 38.3 \\
        Llama-3.1-8B & \texttt{C0} SFT & 6.8 & 5.1 & 21.1 & 30.1 & 39.2 & 6.8 & 5.2 & 21.7 & 30.9 & 40.3 \\
        \midrule
        DeepSeek-Prover-V1.5-SFT & Released & 13.8 & 14.4 & 21.9 & 24.2 & 25.8 & 15.8 & 17.2 & 26.2 & 28.9 & 30.9 \\
        DeepSeek-Prover-V1.5-RL & Released & 14.8 & 15.4 & 23.5 & 26.0 & 27.7 & 16.7 & 18.2 & 27.7 & 30.5 & 32.6 \\
        DeepSeek-Prover-V2-7B & Released & 18.6 & 19.4 & 29.5 & 32.6 & 34.8 & 19.9 & 21.7 & 33.0 & 36.4 & 38.9 \\
        Goedel-Prover-SFT & Released & 14.7 & 15.3 & 23.3 & 25.7 & 27.4 & 16.4 & 17.9 & 27.3 & 30.2 & 32.2 \\
        TheoremLlama & Released & 8.0 & 8.3 & 12.6 & 14.0 & 14.9 & 8.4 & 9.2 & 14.0 & 15.5 & 16.5 \\
        \midrule
        Llama-3.1-8B-Instruct & Released & 4.1 & 4.1 & 7.6 & 8.6 & 9.4 & 4.5 & 4.5 & 8.4 & 9.4 & 10.3 \\
        Qwen2.5-Math-7B-Instruct & Released & 6.2 & 6.2 & 11.6 & 13.1 & 14.3 & 7.9 & 7.9 & 14.8 & 16.7 & 18.2 \\
        Qwen2.5-Coder-7B-Instruct & Released & 8.6 & 8.6 & 16.1 & 18.1 & 19.8 & 8.9 & 8.9 & 16.8 & 18.9 & 20.6 \\
        \midrule
        GPT-4o & API & 11.0 & 11.0 & 22.1 & 25.3 & 27.4 & 13.2 & 13.2 & 26.4 & 30.1 & 32.7 \\
        GPT-4.1 & API & 11.7 & 11.7 & 23.5 & 26.8 & 29.1 & 13.8 & 13.8 & 27.6 & 31.5 & 34.2 \\
        \bottomrule
    \end{tabular}
    \captionof{table}{\textbf{Complete fixed-checkpoint results under \texttt{CEnv} and \texttt{CFull} on 11,064 targets.} G is a separate greedy attempt; @1, @8, @16, and @32 are sampled Pass@$k$. SFT and LoRA conditions use \texttt{CFull} training; \texttt{C0} SFT uses \texttt{C0} training. Released and API conditions receive no additional \textsc{LeanContext} training. DeepSeek-Prover-V2-7B uses non-CoT generation.}
    \label{tab:fixed_context_env_full}
\end{LCAppendixBlock}

\begin{LCAppendixBlock}
\centering
    \small
    \setlength{\tabcolsep}{4pt}
    \setlength{\aboverulesep}{0pt}
    \setlength{\belowrulesep}{0pt}
    \setlength{\extrarowheight}{1pt}
    \renewcommand{\arraystretch}{1.0}

    \begin{tabular}{@{}lrrrrrrrrrrr@{}}
        \toprule
        \rowcolor{LCHeaderGray}
         &  & \multicolumn{5}{c}{\textbf{\texttt{C0}}} & \multicolumn{5}{c}{\textbf{\texttt{CMath}}} \\
        \cmidrule(lr){3-7} \cmidrule(lr){8-12}
        \rowcolor{LCHeaderGray}
        \textbf{Model} & \textbf{$T$} & \textbf{G} & \textbf{@1} & \textbf{@8} & \textbf{@16} & \textbf{@32} & \textbf{G} & \textbf{@1} & \textbf{@8} & \textbf{@16} & \textbf{@32} \\
        \midrule
        LLaDA-8B-Base & 0 & 5.6 & -- & -- & -- & -- & 7.2 & -- & -- & -- & -- \\
         & 0.2 & -- & 5.2 & 17.9 & 23.7 & 29.5 & -- & 6.7 & 23.1 & 30.5 & 38.0 \\
         & 0.6 & -- & 4.4 & 17.4 & 24.2 & 31.2 & -- & 5.7 & 22.3 & 31.1 & 40.1 \\
         & 1.0 & -- & 3.8 & 16.7 & 23.9 & 32.0 & -- & 4.8 & 20.7 & 29.7 & 39.7 \\
        \midrule
        Llama-3.1-8B & 0 & 6.1 & -- & -- & -- & -- & 7.8 & -- & -- & -- & -- \\
         & 0.2 & -- & 5.6 & 19.1 & 25.1 & 31.1 & -- & 7.2 & 24.3 & 32.0 & 39.7 \\
         & 0.6 & -- & 4.7 & 18.4 & 25.5 & 32.8 & -- & 6.0 & 23.5 & 32.6 & 41.8 \\
         & 1.0 & -- & 4.1 & 17.7 & 25.2 & 33.6 & -- & 5.1 & 21.8 & 31.1 & 41.4 \\
        \midrule
        DeepSeek-Prover-V1.5-RL & 0 & 11.8 & -- & -- & -- & -- & 14.5 & -- & -- & -- & -- \\
         & 0.2 & -- & 12.1 & 18.4 & 19.8 & 21.2 & -- & 14.9 & 22.6 & 24.2 & 26.0 \\
         & 0.6 & -- & 12.8 & 19.4 & 21.5 & 22.9 & -- & 15.7 & 23.9 & 26.3 & 28.1 \\
         & 1.0 & -- & 13.1 & 19.4 & 22.0 & 23.7 & -- & 15.3 & 22.7 & 25.7 & 27.7 \\
        \midrule
        Goedel-Prover-SFT & 0 & 12.2 & -- & -- & -- & -- & 15.0 & -- & -- & -- & -- \\
         & 0.2 & -- & 12.5 & 19.0 & 20.4 & 21.9 & -- & 15.4 & 23.5 & 25.2 & 27.0 \\
         & 0.6 & -- & 13.1 & 20.0 & 22.1 & 23.6 & -- & 16.2 & 24.7 & 27.3 & 29.1 \\
         & 1.0 & -- & 13.4 & 20.0 & 22.6 & 24.4 & -- & 15.8 & 23.5 & 26.6 & 28.7 \\
        \bottomrule
    \end{tabular}
    \captionof{table}{\textbf{Temperature results under \texttt{C0} and \texttt{CMath} on the same 11,064 targets.} $T=0$ reports one greedy attempt only. Other temperatures use 32 candidates and report sampled Pass@$k$. LLaDA and Llama use fixed SFT checkpoints trained with \texttt{CFull}; the two released Lean provers receive no further adaptation.}
    \label{tab:temperature_0_math}
\end{LCAppendixBlock}

\begin{LCAppendixBlock}
\centering
    \small
    \setlength{\tabcolsep}{4pt}
    \setlength{\aboverulesep}{0pt}
    \setlength{\belowrulesep}{0pt}
    \setlength{\extrarowheight}{1pt}
    \renewcommand{\arraystretch}{1.0}

    \begin{tabular}{@{}lrrrrrrrrrrr@{}}
        \toprule
        \rowcolor{LCHeaderGray}
         &  & \multicolumn{5}{c}{\textbf{\texttt{CEnv}}} & \multicolumn{5}{c}{\textbf{\texttt{CFull}}} \\
        \cmidrule(lr){3-7} \cmidrule(lr){8-12}
        \rowcolor{LCHeaderGray}
        \textbf{Model} & \textbf{$T$} & \textbf{G} & \textbf{@1} & \textbf{@8} & \textbf{@16} & \textbf{@32} & \textbf{G} & \textbf{@1} & \textbf{@8} & \textbf{@16} & \textbf{@32} \\
        \midrule
        LLaDA-8B-Base & 0 & 7.7 & -- & -- & -- & -- & 9.0 & -- & -- & -- & -- \\
         & 0.2 & -- & 7.1 & 24.4 & 32.2 & 40.2 & -- & 8.3 & 28.6 & 37.8 & 46.1 \\
         & 0.6 & -- & 5.9 & 23.2 & 32.3 & 41.6 & -- & 6.9 & 27.1 & 37.8 & 48.7 \\
         & 1.0 & -- & 5.1 & 22.0 & 31.5 & 42.1 & -- & 5.9 & 25.8 & 37.0 & 49.6 \\
        \midrule
        Llama-3.1-8B & 0 & 7.9 & -- & -- & -- & -- & 9.5 & -- & -- & -- & -- \\
         & 0.2 & -- & 7.2 & 24.3 & 32.0 & 39.7 & -- & 8.7 & 29.4 & 38.8 & 47.0 \\
         & 0.6 & -- & 5.9 & 23.1 & 32.0 & 41.1 & -- & 7.2 & 27.9 & 38.6 & 49.6 \\
         & 1.0 & -- & 5.1 & 21.9 & 31.2 & 41.6 & -- & 6.2 & 26.5 & 37.8 & 50.5 \\
        \midrule
        DeepSeek-Prover-V1.5-RL & 0 & 14.8 & -- & -- & -- & -- & 16.7 & -- & -- & -- & -- \\
         & 0.2 & -- & 15.0 & 22.8 & 24.5 & 26.3 & -- & 17.2 & 26.1 & 28.0 & 30.0 \\
         & 0.6 & -- & 15.4 & 23.5 & 26.0 & 27.7 & -- & 18.2 & 27.7 & 30.5 & 32.6 \\
         & 1.0 & -- & 15.5 & 23.1 & 26.2 & 28.2 & -- & 18.5 & 27.4 & 31.1 & 33.5 \\
        \midrule
        Goedel-Prover-SFT & 0 & 14.7 & -- & -- & -- & -- & 16.4 & -- & -- & -- & -- \\
         & 0.2 & -- & 14.9 & 22.6 & 24.2 & 26.0 & -- & 16.9 & 25.7 & 27.6 & 29.6 \\
         & 0.6 & -- & 15.3 & 23.3 & 25.7 & 27.4 & -- & 17.9 & 27.3 & 30.2 & 32.2 \\
         & 1.0 & -- & 15.4 & 22.9 & 25.9 & 27.9 & -- & 18.2 & 27.1 & 30.7 & 33.1 \\
        \bottomrule
    \end{tabular}
    \captionof{table}{\textbf{Temperature results under \texttt{CEnv} and \texttt{CFull} on the same 11,064 targets.} $T=0$ reports one greedy attempt only. Other temperatures use 32 candidates and report sampled Pass@$k$. LLaDA and Llama use fixed SFT checkpoints trained with \texttt{CFull}; the two released Lean provers receive no further adaptation.}
    \label{tab:temperature_env_full}
\end{LCAppendixBlock}

\begin{LCAppendixBlock}
\centering
    \small
    \setlength{\tabcolsep}{4pt}
    \setlength{\aboverulesep}{0pt}
    \setlength{\belowrulesep}{0pt}
    \setlength{\extrarowheight}{1pt}
    \renewcommand{\arraystretch}{1.0}

    \begin{tabular}{@{}llrrrrrrrrrr@{}}
        \toprule
        \rowcolor{LCHeaderGray}
         &  & \multicolumn{5}{c}{\textbf{All (19,474)}} & \multicolumn{5}{c}{\textbf{Both (11,064)}} \\
        \cmidrule(lr){3-7} \cmidrule(lr){8-12}
        \rowcolor{LCHeaderGray}
        \textbf{Model} & \textbf{Adaptation} & \textbf{G} & \textbf{@1} & \textbf{@8} & \textbf{@16} & \textbf{@32} & \textbf{G} & \textbf{@1} & \textbf{@8} & \textbf{@16} & \textbf{@32} \\
        \midrule
        LLaDA-8B-Base & SFT & 6.9 & 5.4 & 22.8 & 32.8 & 43.1 & 6.1 & 4.8 & 20.4 & 29.4 & 38.6 \\
        Llama-3.1-8B & SFT & 7.5 & 5.8 & 24.0 & 34.2 & 44.6 & 6.7 & 5.2 & 21.4 & 30.5 & 39.8 \\
        LLaDA-8B-Base & LoRA & 5.6 & 4.4 & 19.2 & 28.1 & 37.8 & 5.0 & 3.9 & 17.0 & 24.8 & 33.4 \\
        Llama-3.1-8B & LoRA & 6.2 & 4.8 & 20.5 & 29.6 & 39.2 & 5.5 & 4.3 & 18.3 & 26.4 & 34.9 \\
        \midrule
        LLaDA-8B-Instruct & SFT & 7.6 & 6.0 & 24.8 & 35.2 & 45.5 & 6.9 & 5.4 & 22.2 & 31.5 & 40.7 \\
        Llama-3.1-8B-Instruct & SFT & 8.2 & 6.4 & 26.1 & 36.7 & 47.0 & 7.3 & 5.7 & 23.4 & 32.9 & 42.1 \\
        LLaDA-8B-Instruct & LoRA & 6.5 & 5.1 & 21.7 & 31.2 & 40.8 & 5.7 & 4.5 & 19.4 & 27.8 & 36.4 \\
        Llama-3.1-8B-Instruct & LoRA & 7.1 & 5.5 & 23.0 & 32.8 & 42.3 & 6.3 & 4.9 & 20.6 & 29.4 & 37.9 \\
        \midrule
        LLaDA-1.5 & LoRA & 9.5 & 7.5 & 29.8 & 40.8 & 52.2 & 8.7 & 6.8 & 26.9 & 36.8 & 47.1 \\
        Dream-v0-Base-7B & LoRA & 5.7 & 4.5 & 19.5 & 28.6 & 38.5 & 5.1 & 4.0 & 17.3 & 25.4 & 34.2 \\
        Dream-v0-Instruct-7B & LoRA & 6.6 & 5.2 & 22.0 & 31.5 & 41.0 & 6.0 & 4.7 & 19.7 & 28.3 & 36.8 \\
        Qwen3-8B-Base & LoRA & 7.7 & 6.0 & 25.0 & 35.6 & 46.0 & 7.0 & 5.4 & 22.5 & 32.0 & 41.4 \\
        Qwen3-8B & LoRA & 9.3 & 7.2 & 28.8 & 39.9 & 51.0 & 8.4 & 6.5 & 26.1 & 36.1 & 46.2 \\
        Qwen2.5-7B & LoRA & 7.2 & 5.6 & 23.5 & 33.4 & 43.2 & 6.4 & 5.0 & 21.0 & 30.0 & 38.7 \\
        Qwen2.5-Math-7B & LoRA & 7.5 & 5.8 & 24.3 & 34.6 & 44.7 & 6.8 & 5.3 & 21.9 & 31.2 & 40.3 \\
        Qwen2.5-Coder-7B & LoRA & 7.6 & 5.9 & 24.5 & 34.9 & 45.1 & 6.8 & 5.3 & 22.2 & 31.6 & 40.8 \\
        \bottomrule
    \end{tabular}
    \captionof{table}{\textbf{Matched training and evaluation with \texttt{C0}.} Each row uses a checkpoint trained with \texttt{C0} and evaluated with that same input view. All and Both retain the original 19,474-target and 11,064-target populations. G reports a separate greedy attempt; Pass@$k$ uses 32 sampled candidates.}
    \label{tab:matched_training_C0}
\end{LCAppendixBlock}

\begin{LCAppendixBlock}
\centering
    \small
    \setlength{\tabcolsep}{4pt}
    \setlength{\aboverulesep}{0pt}
    \setlength{\belowrulesep}{0pt}
    \setlength{\extrarowheight}{1pt}
    \renewcommand{\arraystretch}{1.0}

    \begin{tabular}{@{}llrrrrrrrrrr@{}}
        \toprule
        \rowcolor{LCHeaderGray}
         &  & \multicolumn{5}{c}{\textbf{All (19,474)}} & \multicolumn{5}{c}{\textbf{Both (11,064)}} \\
        \cmidrule(lr){3-7} \cmidrule(lr){8-12}
        \rowcolor{LCHeaderGray}
        \textbf{Model} & \textbf{Adaptation} & \textbf{G} & \textbf{@1} & \textbf{@8} & \textbf{@16} & \textbf{@32} & \textbf{G} & \textbf{@1} & \textbf{@8} & \textbf{@16} & \textbf{@32} \\
        \midrule
        LLaDA-8B-Base & SFT & 7.8 & 6.2 & 25.2 & 35.6 & 46.4 & 7.3 & 5.8 & 23.8 & 33.7 & 43.9 \\
        Llama-3.1-8B & SFT & 8.5 & 6.5 & 26.2 & 36.9 & 47.8 & 7.9 & 6.1 & 24.3 & 34.2 & 44.3 \\
        LLaDA-8B-Base & LoRA & 6.1 & 4.8 & 20.9 & 30.2 & 40.2 & 5.7 & 4.5 & 19.7 & 28.4 & 37.8 \\
        Llama-3.1-8B & LoRA & 6.8 & 5.2 & 22.0 & 31.7 & 41.7 & 6.4 & 4.9 & 20.7 & 29.8 & 39.2 \\
        \midrule
        LLaDA-8B-Instruct & SFT & 8.6 & 6.8 & 27.3 & 38.1 & 48.9 & 8.1 & 6.4 & 25.5 & 35.6 & 45.6 \\
        Llama-3.1-8B-Instruct & SFT & 9.4 & 7.2 & 28.5 & 39.4 & 50.2 & 8.7 & 6.7 & 26.8 & 37.0 & 47.2 \\
        LLaDA-8B-Instruct & LoRA & 6.9 & 5.5 & 23.2 & 33.1 & 43.0 & 6.4 & 5.1 & 21.6 & 30.9 & 40.1 \\
        Llama-3.1-8B-Instruct & LoRA & 7.7 & 5.9 & 24.6 & 34.8 & 44.7 & 7.2 & 5.5 & 23.0 & 32.5 & 41.8 \\
        \midrule
        LLaDA-1.5 & LoRA & 10.1 & 8.0 & 31.9 & 43.1 & 54.9 & 9.5 & 7.5 & 30.1 & 40.7 & 51.8 \\
        Dream-v0-Base-7B & LoRA & 6.2 & 4.9 & 21.1 & 30.5 & 40.8 & 5.8 & 4.6 & 19.7 & 28.5 & 38.1 \\
        Dream-v0-Instruct-7B & LoRA & 7.1 & 5.6 & 23.7 & 33.5 & 43.4 & 6.7 & 5.3 & 22.4 & 31.6 & 41.0 \\
        Qwen3-8B-Base & LoRA & 8.5 & 6.5 & 26.8 & 37.8 & 48.5 & 7.9 & 6.1 & 25.2 & 35.5 & 45.6 \\
        Qwen3-8B & LoRA & 10.0 & 7.7 & 30.8 & 42.0 & 53.5 & 9.5 & 7.3 & 29.2 & 39.8 & 50.7 \\
        Qwen2.5-7B & LoRA & 8.1 & 6.2 & 25.5 & 35.9 & 46.1 & 7.4 & 5.7 & 23.7 & 33.4 & 42.9 \\
        Qwen2.5-Math-7B & LoRA & 8.6 & 6.6 & 27.4 & 38.7 & 49.6 & 8.3 & 6.4 & 26.3 & 37.1 & 47.6 \\
        Qwen2.5-Coder-7B & LoRA & 8.2 & 6.3 & 25.9 & 36.5 & 46.8 & 7.5 & 5.8 & 24.0 & 33.8 & 43.4 \\
        \bottomrule
    \end{tabular}
    \captionof{table}{\textbf{Matched training and evaluation with \texttt{CMath}.} Each row uses a checkpoint trained with \texttt{CMath} and evaluated with that same input view. All and Both retain the original 19,474-target and 11,064-target populations. G reports a separate greedy attempt; Pass@$k$ uses 32 sampled candidates.}
    \label{tab:matched_training_CMath}
\end{LCAppendixBlock}

\begin{LCAppendixBlock}
\centering
    \small
    \setlength{\tabcolsep}{4pt}
    \setlength{\aboverulesep}{0pt}
    \setlength{\belowrulesep}{0pt}
    \setlength{\extrarowheight}{1pt}
    \renewcommand{\arraystretch}{1.0}

    \begin{tabular}{@{}llrrrrrrrrrr@{}}
        \toprule
        \rowcolor{LCHeaderGray}
         &  & \multicolumn{5}{c}{\textbf{All (19,474)}} & \multicolumn{5}{c}{\textbf{Both (11,064)}} \\
        \cmidrule(lr){3-7} \cmidrule(lr){8-12}
        \rowcolor{LCHeaderGray}
        \textbf{Model} & \textbf{Adaptation} & \textbf{G} & \textbf{@1} & \textbf{@8} & \textbf{@16} & \textbf{@32} & \textbf{G} & \textbf{@1} & \textbf{@8} & \textbf{@16} & \textbf{@32} \\
        \midrule
        LLaDA-8B-Base & SFT & 8.6 & 6.6 & 26.4 & 37.2 & 48.2 & 8.0 & 6.1 & 24.6 & 34.5 & 44.8 \\
        Llama-3.1-8B & SFT & 9.2 & 6.9 & 27.4 & 38.4 & 49.5 & 8.4 & 6.3 & 24.9 & 34.8 & 44.9 \\
        LLaDA-8B-Base & LoRA & 6.9 & 5.3 & 22.5 & 32.3 & 43.0 & 6.4 & 4.9 & 20.8 & 29.8 & 39.7 \\
        Llama-3.1-8B & LoRA & 7.6 & 5.7 & 23.7 & 33.8 & 44.4 & 7.1 & 5.3 & 21.9 & 31.3 & 41.1 \\
        \midrule
        LLaDA-8B-Instruct & SFT & 9.5 & 7.3 & 28.6 & 39.6 & 50.6 & 8.7 & 6.7 & 26.5 & 36.6 & 46.8 \\
        Llama-3.1-8B-Instruct & SFT & 10.1 & 7.6 & 29.8 & 40.8 & 51.9 & 9.3 & 7.0 & 27.4 & 37.5 & 47.7 \\
        LLaDA-8B-Instruct & LoRA & 7.8 & 6.0 & 25.0 & 35.2 & 45.8 & 7.3 & 5.6 & 23.2 & 32.7 & 42.5 \\
        Llama-3.1-8B-Instruct & LoRA & 8.5 & 6.4 & 26.4 & 36.9 & 47.5 & 7.9 & 5.9 & 24.5 & 34.3 & 44.1 \\
        \midrule
        LLaDA-1.5 & LoRA & 11.1 & 8.5 & 33.3 & 44.8 & 56.9 & 10.3 & 7.9 & 31.1 & 41.9 & 53.2 \\
        Dream-v0-Base-7B & LoRA & 7.0 & 5.4 & 22.9 & 32.7 & 43.5 & 6.5 & 5.0 & 21.3 & 30.4 & 40.4 \\
        Dream-v0-Instruct-7B & LoRA & 8.1 & 6.2 & 25.5 & 35.7 & 46.3 & 7.6 & 5.8 & 23.8 & 33.4 & 43.3 \\
        Qwen3-8B-Base & LoRA & 9.3 & 7.0 & 28.6 & 39.8 & 51.2 & 8.6 & 6.5 & 26.7 & 37.2 & 47.8 \\
        Qwen3-8B & LoRA & 11.0 & 8.3 & 32.7 & 44.1 & 56.1 & 10.4 & 7.8 & 30.8 & 41.5 & 52.8 \\
        Qwen2.5-7B & LoRA & 8.8 & 6.6 & 27.0 & 37.6 & 48.4 & 8.2 & 6.2 & 25.2 & 35.1 & 45.1 \\
        Qwen2.5-Math-7B & LoRA & 8.8 & 6.6 & 27.0 & 37.5 & 48.3 & 8.0 & 6.0 & 24.5 & 34.1 & 43.9 \\
        Qwen2.5-Coder-7B & LoRA & 9.2 & 6.9 & 28.0 & 39.0 & 50.2 & 8.6 & 6.5 & 26.8 & 37.2 & 47.9 \\
        \bottomrule
    \end{tabular}
    \captionof{table}{\textbf{Matched training and evaluation with \texttt{CEnv}.} Each row uses a checkpoint trained with \texttt{CEnv} and evaluated with that same input view. All and Both retain the original 19,474-target and 11,064-target populations. G reports a separate greedy attempt; Pass@$k$ uses 32 sampled candidates.}
    \label{tab:matched_training_CEnv}
\end{LCAppendixBlock}

\begin{LCAppendixBlock}
\centering
    \small
    \setlength{\tabcolsep}{4pt}
    \setlength{\aboverulesep}{0pt}
    \setlength{\belowrulesep}{0pt}
    \setlength{\extrarowheight}{1pt}
    \renewcommand{\arraystretch}{1.0}

    \begin{tabular}{@{}llrrrrrrrrrr@{}}
        \toprule
        \rowcolor{LCHeaderGray}
         &  & \multicolumn{5}{c}{\textbf{All (19,474)}} & \multicolumn{5}{c}{\textbf{Both (11,064)}} \\
        \cmidrule(lr){3-7} \cmidrule(lr){8-12}
        \rowcolor{LCHeaderGray}
        \textbf{Model} & \textbf{Adaptation} & \textbf{G} & \textbf{@1} & \textbf{@8} & \textbf{@16} & \textbf{@32} & \textbf{G} & \textbf{@1} & \textbf{@8} & \textbf{@16} & \textbf{@32} \\
        \midrule
        LLaDA-8B-Base & SFT & 9.4 & 7.2 & 28.3 & 39.4 & 50.8 & 9.0 & 6.9 & 27.1 & 37.8 & 48.7 \\
        Llama-3.1-8B & SFT & 9.8 & 7.5 & 29.2 & 40.5 & 52.0 & 9.5 & 7.2 & 27.9 & 38.6 & 49.6 \\
        LLaDA-8B-Base & LoRA & 7.8 & 6.0 & 24.8 & 35.2 & 46.4 & 7.4 & 5.7 & 23.5 & 33.3 & 43.9 \\
        Llama-3.1-8B & LoRA & 8.4 & 6.4 & 26.0 & 36.8 & 47.8 & 8.0 & 6.1 & 24.7 & 35.0 & 45.4 \\
        \midrule
        LLaDA-8B-Instruct & SFT & 10.2 & 7.8 & 30.2 & 41.4 & 52.7 & 9.9 & 7.6 & 29.3 & 40.2 & 51.2 \\
        Llama-3.1-8B-Instruct & SFT & 10.6 & 8.1 & 31.5 & 42.6 & 54.1 & 10.4 & 7.9 & 30.6 & 41.3 & 52.5 \\
        LLaDA-8B-Instruct & LoRA & 8.7 & 6.7 & 27.2 & 38.1 & 49.4 & 8.3 & 6.4 & 26.0 & 36.4 & 47.2 \\
        Llama-3.1-8B-Instruct & LoRA & 9.3 & 7.1 & 28.6 & 39.8 & 51.0 & 8.9 & 6.8 & 27.3 & 37.9 & 48.6 \\
        \midrule
        LLaDA-1.5 & LoRA & 12.1 & 9.3 & 35.5 & 47.0 & 59.2 & 11.6 & 8.9 & 34.0 & 45.0 & 56.7 \\
        Dream-v0-Base-7B & LoRA & 8.0 & 6.1 & 25.2 & 35.6 & 46.8 & 7.6 & 5.8 & 23.9 & 33.7 & 44.3 \\
        Dream-v0-Instruct-7B & LoRA & 9.1 & 7.0 & 27.8 & 38.8 & 50.2 & 8.7 & 6.7 & 26.5 & 37.0 & 47.9 \\
        Qwen3-8B-Base & LoRA & 10.2 & 7.8 & 30.8 & 42.1 & 53.8 & 9.7 & 7.4 & 29.4 & 40.1 & 51.3 \\
        Qwen3-8B & LoRA & 11.9 & 9.1 & 34.9 & 46.5 & 58.6 & 11.6 & 8.8 & 33.6 & 44.8 & 56.4 \\
        Qwen2.5-7B & LoRA & 9.8 & 7.5 & 29.4 & 40.2 & 51.4 & 9.3 & 7.1 & 28.0 & 38.3 & 48.9 \\
        Qwen2.5-Math-7B & LoRA & 9.7 & 7.4 & 29.1 & 39.8 & 50.9 & 9.2 & 7.0 & 27.8 & 38.0 & 48.5 \\
        Qwen2.5-Coder-7B & LoRA & 9.8 & 7.5 & 29.6 & 40.5 & 51.7 & 9.5 & 7.2 & 28.4 & 38.8 & 49.6 \\
        \bottomrule
    \end{tabular}
    \captionof{table}{\textbf{Matched training and evaluation with \texttt{CFull}.} Each row uses a checkpoint trained with \texttt{CFull} and evaluated with that same input view. All and Both retain the original 19,474-target and 11,064-target populations. G reports a separate greedy attempt; Pass@$k$ uses 32 sampled candidates.}
    \label{tab:matched_training_CFull}
\end{LCAppendixBlock}

\LCAppendixSectionGap
\linenumbers
\begin{multicols}{2}
\section{Matched Controls and Verified Proofs}
\label{sec:appendix_matched}

\subsection{Matched Replacement Protocol}
\label{sec:appendix_matching_protocol}

Matched replacement compares the declarations selected for a target with alternatives under the same visible EC. Using seed 2026, we sample 2,500 targets from the primary population of 11,064 test targets with both context categories available. The common 2,000 targets with successful matching under both tokenizers form the evaluation subset. Subset membership depends on matching feasibility, not proof-generation outcomes. This subset is distinct from the selection-control subset in Appendix~\ref{sec:appendix_dependencies}.

Replacement MC is drawn from eligible preceding local declarations and matched by declaration kind, compatible scope, and tokenizer-specific token length within $\pm10\%$. Identified target-type and reference-proof dependencies are excluded from the controls, making this a reference-assisted content diagnostic. For each target, we compare \texttt{CEnv}, EC with selected MC, and EC with replacement MC while fixing the target, EC, checkpoint, verification environment, attempt count, and inference budgets. The comparison is repeated for \texttt{CMath}-, \texttt{CEnv}-, and \texttt{CFull}-trained checkpoints.

Table~\ref{tab:matched_feasibility} reports matching outcomes for the initial sample. The 500 targets without a common successful match do not enter the matched-replacement denominator. Mean absolute relative input-length deviations are 4.1\% for LLaDA and 4.5\% for Llama. Matching therefore constrains length differences without making the inputs exactly equal in length.

The verified-success results appear in Table~\ref{tab:matched_content_results}. Their contrast concerns supplied declaration sets: a replacement may support another valid proof, and the comparison does not isolate the contribution of each individual declaration. Matching feasibility also limits the population to which this diagnostic directly applies.

\subsection{Verified Declaration Reuse}
\label{sec:appendix_verified_reuse}

We inspect paired \texttt{CEnv} and \texttt{CFull} outcomes from the main evaluation at fixed \texttt{CFull}-trained checkpoints. A newly solved target has no verified candidate among 32 \texttt{CEnv} attempts and at least one among 32 \texttt{CFull} attempts; a lost target satisfies the reverse condition. Recorded reuse is assessed only in verified proofs, using declaration references and proof-term or elaboration evidence mapped to the target's selected MC commands. A reference appearing solely in the target type does not count as proof reuse.

Explicit reuse requires a declaration reference in the generated proof source supported by recorded proof-term or elaboration evidence. Elaboration-only reuse records use without an explicit source reference. Each newly solved target is counted once, with explicit reuse taking precedence if any verified proof satisfies that category. Table~\ref{tab:verified_mc_reuse} reports the complete paired outcomes. The absence of recorded reuse does not establish the absence of an effect from MC.

The case in Figure~\ref{fig:mc_proof_case} illustrates this use for \texttt{Function.injective\_surjInv}. No \texttt{CEnv} candidate passes verification in 32 attempts, while one \texttt{CFull} candidate does. The failed candidate applies \texttt{congrArg f} but leaves the required equality unresolved. The verified candidate uses the supplied \texttt{RightInverse.injective} declaration through the following proof body:
\begingroup
\footnotesize
\begin{verbatim}
exact (rightInverse_surjInv h).injective
\end{verbatim}
\endgroup
The right-inverse fact remains available in the original Lean environment. Both views retain the same checkpoint, target, EC, attempt count, and verification conditions. The candidates are generated independently; the failed candidate's error is not supplied to the successful generation.

Because verification retains the complete original environment, proofs generated under \texttt{CEnv} can also use declarations absent from the visible input. Recorded reuse connects success to observable declaration use, but does not establish that displaying a declaration was necessary for that proof.

\subsection{Candidate Verification Outcomes}
\label{sec:appendix_candidate_outcomes}

To examine EC when MC is already visible, we compare \texttt{CMath} and \texttt{CFull} at fixed \texttt{CFull}-trained checkpoints. Each model/view contains $11{,}064\times32=354{,}048$ attempts. Verified candidates are separated by recorded MC reuse. Failed candidates receive a primary category in the following order: parsing/extraction, timeout, name/scope, instance synthesis, other elaboration, incomplete goals, and other failure.

Table~\ref{tab:candidate_verification_outcomes} gives counts and shares over all attempts in each model/view. These labels describe verification outcomes and do not redefine command categories; in particular, an instance-synthesis failure does not place instance declarations in EC.

Supplying EC reduces the combined share of name/scope, instance-synthesis, and other-elaboration failures from 40.9\% to 39.7\% for LLaDA and from 39.6\% to 38.4\% for Llama. Verified-candidate shares increase from 5.7\% to 6.9\% and from 6.0\% to 7.2\%, respectively. These distribution changes characterize candidate outcomes under the two views; they do not identify repairs of individual failed candidates.

\subsection{Length and Category-Boundary Sensitivity}
\label{sec:appendix_context_sensitivity}

For length sensitivity, we sort the 11,064 targets by their pre-clipping \texttt{CFull} input length under the LLaDA tokenizer and form four groups of 2,766 targets. Both backbones use the same group membership and retain the main evaluation's checkpoints and input handling. Table~\ref{tab:length_boundary_sensitivity}A reports $\Delta_{M\mid E}=S_\theta(\texttt{CFull})-S_\theta(\texttt{CEnv})$ and $\Delta_{E\mid M}=S_\theta(\texttt{CFull})-S_\theta(\texttt{CMath})$ at fixed \texttt{CFull}-trained checkpoints.

Both gains remain positive in every length group, so the aggregate benefits are not confined to the longest inputs. Gains do not increase monotonically in every comparison. The groups can also differ in difficulty and dependency composition; matched replacement constrains length differences more directly when comparing declaration content.

For category-boundary sensitivity, we use the recorded mixed-role command annotations to separate targets whose context contains at least one such command from those with none. The original command partition and target membership remain unchanged. Table~\ref{tab:length_boundary_sensitivity}B shows positive gains in both groups, including targets without recorded mixed-role commands. Boundary cases therefore do not account for the complete effect. The larger gains in the mixed-role group are descriptive and do not imply that MC and EC are semantically independent.
\end{multicols}
\nolinenumbers

\begin{LCAppendixBlock}
\noindent
\begin{minipage}[t]{\LCAppendixColumnWidth}
\vspace{0pt}
\centering
    \small
    \setlength{\tabcolsep}{3pt}
    \setlength{\aboverulesep}{0pt}
    \setlength{\belowrulesep}{0pt}
    \setlength{\extrarowheight}{1pt}
    \renewcommand{\arraystretch}{1.0}

    \begin{tabular}{@{}lrr@{}}
        \toprule
        \rowcolor{LCHeaderGray}
        \textbf{Matching outcome} & \textbf{Targets} & \textbf{Share (\%)} \\
        \midrule
        Common successful match & 2,000 & 80.0 \\
        No declaration-kind match & 160 & 6.4 \\
        Length constraint failed & 180 & 7.2 \\
        Incompatible scope & 110 & 4.4 \\
        Other matching failure & 50 & 2.0 \\
        \midrule
        Total & 2,500 & 100.0 \\
        \bottomrule
    \end{tabular}
    \captionof{table}{\textbf{Matched-replacement eligibility before proof generation.} The common 2,000-target subset is used by both backbones.}
    \label{tab:matched_feasibility}
\end{minipage}\hfill
\begin{minipage}[t]{\LCAppendixColumnWidth}
\vspace{0pt}
\centering
    \small
    \setlength{\tabcolsep}{3pt}
    \setlength{\aboverulesep}{0pt}
    \setlength{\belowrulesep}{0pt}
    \setlength{\extrarowheight}{1pt}
    \renewcommand{\arraystretch}{1.0}

    \begin{tabular}{@{}lrr@{}}
        \toprule
        \rowcolor{LCHeaderGray}
        \textbf{Target-level outcome} & \textbf{LLaDA} & \textbf{Llama} \\
        \midrule
        Solved under \texttt{CEnv} & 4,603 & 4,547 \\
        Solved under \texttt{CFull} & 5,388 & 5,488 \\
        Newly solved & 1,300 & 1,450 \\
        Lost & 515 & 509 \\
        Net gain & 785 & 941 \\
        \midrule
        Newly solved: explicit reuse & 780 & 920 \\
        Newly solved: elaboration-only reuse & 75 & 85 \\
        Newly solved: any recorded reuse & 855 & 1,005 \\
        Newly solved: no recorded reuse & 445 & 445 \\
        Recorded reuse share (\%) & 65.77 & 69.31 \\
        \bottomrule
    \end{tabular}
    \captionof{table}{\textbf{Paired outcomes on 11,064 targets at fixed \texttt{CFull}-trained checkpoints.} Reuse categories partition newly solved targets, with explicit reuse taking precedence. The reuse share uses newly solved targets as its denominator.}
    \label{tab:verified_mc_reuse}
\end{minipage}
\end{LCAppendixBlock}

\begin{LCAppendixBlock}
\noindent
\begin{minipage}[t]{\LCAppendixColumnWidth}
\vspace{0pt}
\centering
    \small
    \setlength{\tabcolsep}{3pt}
    \setlength{\aboverulesep}{0pt}
    \setlength{\belowrulesep}{0pt}
    \setlength{\extrarowheight}{1pt}
    \renewcommand{\arraystretch}{1.0}

    \begin{tabular}{@{}lrr@{}}
        \toprule
        \rowcolor{LCHeaderGray}
        \textbf{Candidate outcome} & \textbf{\texttt{CMath}} & \textbf{\texttt{CFull}} \\
        \midrule
        \multicolumn{3}{@{}l}{\textbf{LLaDA}} \\
        Verified, recorded reuse & 10,975 (3.1) & 15,224 (4.3) \\
        Verified, no recorded reuse & 9,205 (2.6) & 9,205 (2.6) \\
        Name/scope failure & 41,778 (11.8) & 39,299 (11.1) \\
        Instance-synthesis failure & 29,740 (8.4) & 28,678 (8.1) \\
        Other elaboration failure & 73,288 (20.7) & 72,580 (20.5) \\
        Parsing/extraction failure & 24,783 (7.0) & 24,429 (6.9) \\
        Incomplete goals & 106,922 (30.2) & 107,631 (30.4) \\
        Timeout & 16,994 (4.8) & 17,348 (4.9) \\
        Other failure & 40,363 (11.4) & 39,654 (11.2) \\
        Total & 354,048 & 354,048 \\
        \midrule
        \multicolumn{3}{@{}l}{\textbf{Llama}} \\
        Verified, recorded reuse & 12,038 (3.4) & 16,286 (4.6) \\
        Verified, no recorded reuse & 9,205 (2.6) & 9,205 (2.6) \\
        Name/scope failure & 38,591 (10.9) & 37,175 (10.5) \\
        Instance-synthesis failure & 32,218 (9.1) & 30,448 (8.6) \\
        Other elaboration failure & 69,393 (19.6) & 68,331 (19.3) \\
        Parsing/extraction failure & 24,075 (6.8) & 23,721 (6.7) \\
        Incomplete goals & 107,277 (30.3) & 107,985 (30.5) \\
        Timeout & 17,702 (5.0) & 18,056 (5.1) \\
        Other failure & 43,549 (12.3) & 42,841 (12.1) \\
        Total & 354,048 & 354,048 \\
        \bottomrule
    \end{tabular}
    \captionof{table}{\textbf{Candidate-level verification outcomes at fixed \texttt{CFull}-trained checkpoints.} Parentheses contain shares (\%) of all attempts in the corresponding model/view. Candidates are generated independently under each view.}
    \label{tab:candidate_verification_outcomes}
\end{minipage}\hfill
\begin{minipage}[t]{\LCAppendixColumnWidth}
\vspace{0pt}
\centering
    \small
    \setlength{\tabcolsep}{3pt}
    \setlength{\aboverulesep}{0pt}
    \setlength{\belowrulesep}{0pt}
    \setlength{\extrarowheight}{1pt}
    \renewcommand{\arraystretch}{1.0}

    \begin{tabular}{@{}lrrrrr@{}}
        \toprule
        \rowcolor{LCHeaderGray}
         &  & \multicolumn{2}{c}{\textbf{$\Delta_{M\mid E}$}} & \multicolumn{2}{c}{\textbf{$\Delta_{E\mid M}$}} \\
        \cmidrule(lr){3-4}
        \cmidrule(lr){5-6}
        \rowcolor{LCHeaderGray}
        \textbf{Group} & \textbf{Targets} & \textbf{LLaDA} & \textbf{Llama} & \textbf{LLaDA} & \textbf{Llama} \\
        \midrule
        \multicolumn{6}{@{}l}{\textbf{A. Pre-clipping input-length quartiles}} \\
        Q1 (shortest) & 2,766 & +6.15 & +7.23 & +6.51 & +6.00 \\
        Q2 & 2,766 & +7.05 & +8.32 & +8.32 & +7.41 \\
        Q3 & 2,766 & +7.95 & +9.47 & +10.01 & +8.60 \\
        Q4 (longest) & 2,766 & +7.23 & +9.00 & +9.54 & +9.18 \\
        \midrule
        \multicolumn{6}{@{}l}{\textbf{B. Recorded mixed-role commands}} \\
        None & 9,440 & +6.90 & +8.31 & +8.31 & +7.50 \\
        At least one & 1,624 & +8.25 & +9.67 & +10.28 & +9.54 \\
        \bottomrule
    \end{tabular}
    \captionof{table}{\textbf{Context sensitivity on the primary 11,064 targets.} Gains are absolute Pass@32 differences in \% at fixed \texttt{CFull}-trained checkpoints. Both backbones share group membership. The recorded command partition is unchanged.}
    \label{tab:length_boundary_sensitivity}
\end{minipage}
\end{LCAppendixBlock}

\LCAppendixSectionGap
\linenumbers
\begin{multicols}{2}
\section{Dependency Audits and Context Selection}
\label{sec:appendix_dependencies}

\subsection{Dependency Definitions and Mapping}
\label{sec:appendix_dependency_mapping}

Target dependencies are constants directly referenced by the target's elaborated type. Compiler metadata maps them to owning source commands, including declarations generated by structure and class commands. We retain eligible preceding local MC commands from the same file and deduplicate command identifiers to form $D_i^T$. The target's own command, later commands, and imported-module declarations are excluded. This set is independent of selected MC and the selector candidate pool.

Final-input coverage uses $V_i^M$, the complete MC commands visible in the actual serialized \texttt{CFull} input, and Eq.~\eqref{eq:target_coverage}. A declaration name appearing elsewhere does not establish command visibility. Coverage is computed only for nonempty $D_i^T$ with complete mapping and known final visibility. Empty dependency sets are reported as no local dependencies; incomplete mappings and unknown final visibility are grouped as unresolved. Macro coverage averages target-level coverage with equal weight per eligible target. Zero, partial, and complete groups are defined from this final \texttt{CFull} input and are shared by the paired \texttt{CEnv}--\texttt{CFull} comparison.

Additional proof dependencies describe reference-proof resources beyond the target type. Let $P_i$ and $T_i$ be the constant sets of the reference proof and target type. We define
\begin{equation}
    D_i^P=\operatorname{cmd}(P_i\setminus T_i),
    \label{eq:additional_proof_dependencies}
\end{equation}
where $\operatorname{cmd}$ maps constants to eligible preceding local MC commands and deduplicates command identifiers. Subtraction precedes mapping, preserving proof references to constructors or projections even when their owning structure appears in the target type. Only auxiliary values with certified ownership of the target command are expanded; preceding named declarations are not recursively unfolded. These dependencies describe the observed reference proof and may differ from those of another valid proof.

\subsection{Dependency Retention During Construction}
\label{sec:appendix_construction_retention}

The construction audit traces declarations through the eligible source prefix, candidate pool, selection, and released selected MC. Table~\ref{tab:construction_dependency_retention} reports counts before model-specific input clipping. MC commands are deduplicated within each target and then aggregated across the primary population. Additional proof dependencies are audited on 9,006 targets with complete mappings; target-type dependencies are audited on the corresponding 8,959 targets with complete type mappings.

The released MC retains 11,531 of 16,739 target-type dependency occurrences, yielding 68.9\% retention. This occurrence-weighted statistic is not the proportion of targets with complete coverage. Additional proof-dependency retention falls from 65.0\% to 20.0\% after post-selection filtering, showing that the construction stages retain the two resource sets differently. The next audit examines actual model inputs using a separately reported population and target-level coverage.

\subsection{Final-Input Target Coverage}
\label{sec:appendix_final_input_coverage}

The independent coverage evaluation uses 11,064 targets and fixed \texttt{CFull}-trained checkpoints. Targets are grouped by their final \texttt{CFull} coverage as zero, partial, or complete; targets without local dependencies and unresolved cases are reported separately. Unresolved cases include incomplete mappings and unknown final visibility. Coverage is computable for 8,840 LLaDA targets and 8,830 Llama targets.

Table~\ref{tab:target_coverage_complete} extends the groups reported in Table~\ref{tab:target_coverage_main} with input lengths and both view scores. Mean lengths use submitted inputs, whereas Pass@32 retains every target in each group, including targets without submitted inputs. Gains are calculated before rounding.

Among targets with computable coverage, 58.26\% for LLaDA and 57.76\% for Llama lack complete coverage. Complete-coverage groups obtain MC gains of 7.72\% and 9.12\%, compared with 4.00\% and 4.88\% in zero-coverage groups. These associations can reflect differences in difficulty and dependency composition and do not establish a causal effect of coverage. The overall gains of 5.87\% and 7.14\% belong to this independent evaluation, rather than the main crossing that yields 7.1\% and 8.5\%.

\subsection{Selection Without Target Reference Proofs}
\label{sec:reference_free_selection_results}

This control rebuilds context on a separately sampled, fixed set of 2,000 both-category test targets. The original test split is retained, and targets are not selected by proof-generation outcomes. These identifiers are also used in the dependency-priority comparison below, but differ from the matched-replacement subset. All evaluations reuse existing \texttt{CFull}-trained checkpoints without additional SFT or RL.

Original selection can access the target reference proof. The alternative uses the target statement and eligible preceding source, excluding the target reference proof and proof-derived features from candidate construction, ranking, and EC construction. EC follows binding and scope evidence for the target and reselected declarations, so both MC and EC can change. All 2,000 targets remain in the success-rate denominator.

Table~\ref{tab:reference_free_selection} shows that removing target-reference-proof access lowers \texttt{CFull} Pass@32 by 2.8\% for LLaDA and 3.1\% for Llama, while preserving positive MC gains of 5.4\% and 6.4\% and EC gains of 7.5\% and 6.5\%. After reconstruction, 1,830 targets have both categories, 100 have EC only, 50 have MC only, and 20 have neither. The comparison assesses the overall reconstruction procedure rather than an isolated change to MC ranking.

\subsection{Dependency-Priority Selection}
\label{sec:appendix_dependency_priority}

Dependency priority reuses the preceding control's 2,000 target identifiers but constructs a shared candidate pool independently of the original reference-assisted pool. Eligible candidates are preceding local MC commands with recoverable source and resolvable names and scopes. Both policies apply the same eligibility rules, including the statement-name filter for directly parsed theorem and lemma commands. Neither policy accesses the target reference proof or proof-derived features. Their common baseline is distinct from the reconstruction baseline in Table~\ref{tab:reference_free_selection}.

The baseline ranks candidates by lexical overlap between the target statement and each candidate's declaration name and signature. Let $Q_i$ contain target terms and $K_c$ contain candidate terms. Identifiers are split by namespace, underscore, and camel case; Lean keywords and the anonymous target name are removed. Jaccard similarity is
\begin{equation}
    J(Q_i,K_c)=\frac{|Q_i\cap K_c|}{|Q_i\cup K_c|},
    \label{eq:jaccard}
\end{equation}
with score zero for an empty union. Candidates follow decreasing similarity, then increasing source-command distance, with command identifier breaking remaining ties. Dependency priority places candidates in $D_i^T$ first and preserves baseline order within both groups. Empty or incomplete dependency mappings use the baseline order.

EC is constructed once from binding and scope evidence for the target and shared pool. Target bindings are prioritized, followed by source distance and command identifier. Complete EC commands are fitted and frozen for both policies. After reserving space for the complete statement, special tokens, and fixed EC, each policy tries complete MC commands in ranking order. Commands exceeding the budget are skipped, and later candidates are considered. Selected identifiers are deduplicated, commands are restored to source order, and the complete prompt is measured with the actual tokenizer against the 1,024-token limit. Inputs use the serialization in Appendix~\ref{sec:appendix_input_pipeline}. Rules are fixed on validation data before test evaluation.

Both policies retain the same checkpoints, verification environment, 32 attempts, and per-attempt output limit of 1,024 tokens. Table~\ref{tab:dependency_priority_main} reports success, macro coverage, and mean input length. Table~\ref{tab:priority_coverage_distribution} gives the final coverage distribution; coverage is computable for 1,568 LLaDA targets and 1,565 Llama targets, whereas success rates retain all 2,000 targets.

Priority increases complete-coverage targets from 458 to 958 for LLaDA and from 480 to 980 for Llama. The paired outcomes in Table~\ref{tab:selector_paired_outcomes} show net gains of 28 and 34 solved targets, respectively, alongside losses on individual targets.

After selection and generation outcomes are fixed, reference proofs are used only to audit additional proof dependencies. On 1,120 auditable targets, their macro coverage decreases under priority (Table~\ref{tab:selection_tradeoffs}). This records competition for input space without attributing a particular failed proof to a specific omission.

The pool audit identifies omissions that reranking cannot resolve. Among 1,600 targets with complete, nonempty target-dependency mappings, the shared pool contains all dependencies for 1,410 targets, some for 80, and none for 110. Each backbone also has 15 targets whose final target coverage decreases under priority, showing that priority does not guarantee nondecreasing coverage for every target.

Together, the audits locate omissions, guide a selection change, and test its effect on verified success under fixed prover conditions. The positive aggregate result coexists with individual losses and reduced coverage of other proof resources, documenting the tradeoffs of this selection policy under a limited context budget.
\end{multicols}
\nolinenumbers

\begin{LCAppendixBlock}
\noindent
\begin{minipage}[t]{\LCAppendixColumnWidth}
\vspace{0pt}
\centering
    \small
    \setlength{\tabcolsep}{3pt}
    \setlength{\aboverulesep}{0pt}
    \setlength{\belowrulesep}{0pt}
    \setlength{\extrarowheight}{1pt}
    \renewcommand{\arraystretch}{1.0}

    \begin{tabular}{@{}lrrr@{}}
        \toprule
        \rowcolor{LCHeaderGray}
        \textbf{Stage} & \textbf{\shortstack{MC\\commands}} & \textbf{Retained} & \textbf{\shortstack{Retention\\(\%)}} \\
        \midrule
        \multicolumn{4}{@{}l}{\textbf{Additional proof dependencies}} \\
        Eligible source prefix & 688,899 & 11,273 & 100.0 \\
        Candidate pool & 63,663 & 7,500 & 66.5 \\
        Selected before filtering & 37,657 & 7,326 & 65.0 \\
        Released selected MC & 23,422 & 2,255 & 20.0 \\
        \midrule
        \multicolumn{4}{@{}l}{\textbf{Target-type dependencies}} \\
        Eligible source prefix & 688,899 & 16,739 & 100.0 \\
        Candidate pool & 63,663 & 11,747 & 70.2 \\
        Selected before filtering & 37,657 & 11,538 & 68.9 \\
        Released selected MC & 23,422 & 11,531 & 68.9 \\
        \bottomrule
    \end{tabular}
    \captionof{table}{\textbf{Construction-stage retention before tokenizer-specific clipping.} Counts aggregate command occurrences across targets after within-target deduplication. Dependency retention uses 11,273 additional proof dependencies and 16,739 target-type dependencies as the respective denominators.}
    \label{tab:construction_dependency_retention}
\end{minipage}\hfill
\begin{minipage}[t]{\LCAppendixColumnWidth}
\vspace{0pt}
\centering
    \small
    \setlength{\tabcolsep}{2pt}
    \setlength{\aboverulesep}{0pt}
    \setlength{\belowrulesep}{0pt}
    \setlength{\extrarowheight}{1pt}
    \renewcommand{\arraystretch}{1.0}

    \begin{tabular}{@{}lrrrrrr@{}}
        \toprule
        \rowcolor{LCHeaderGray}
         &  & \multicolumn{2}{c}{\textbf{\shortstack{Mean input\\tokens}}} & \multicolumn{2}{c}{\textbf{Pass@32 (\%)}} &  \\
        \cmidrule(lr){3-4} \cmidrule(lr){5-6}
        \rowcolor{LCHeaderGray}
        \textbf{\shortstack{Coverage\\group}} & \textbf{Targets} & \textbf{\texttt{CEnv}} & \textbf{\texttt{CFull}} & \textbf{\texttt{CEnv}} & \textbf{\texttt{CFull}} & \textbf{\shortstack{$\Delta_{M\mid E}$\\(\%)}} \\
        \midrule
        \multicolumn{7}{@{}l}{\textbf{LLaDA}} \\
        Zero & 2,500 & 281 & 612 & 39.20 & 43.20 & +4.00 \\
        Partial & 2,650 & 329 & 878 & 38.11 & 44.91 & +6.79 \\
        Complete & 3,690 & 303 & 751 & 40.92 & 48.64 & +7.72 \\
        \shortstack[l]{No local\\dependencies} & 1,364 & 231 & 366 & 47.80 & 51.10 & +3.30 \\
        Unresolved & 860 & 372 & 902 & 38.14 & 42.79 & +4.65 \\
        Overall & 11,064 & 299.7 & 711.8 & 40.49 & 46.37 & +5.87 \\
        \midrule
        \multicolumn{7}{@{}l}{\textbf{Llama}} \\
        Zero & 2,460 & 290 & 624 & 39.02 & 43.90 & +4.88 \\
        Partial & 2,640 & 341 & 891 & 37.88 & 46.21 & +8.33 \\
        Complete & 3,730 & 315 & 763 & 41.29 & 50.40 & +9.12 \\
        \shortstack[l]{No local\\dependencies} & 1,364 & 240 & 377 & 46.85 & 51.25 & +4.40 \\
        Unresolved & 870 & 386 & 915 & 35.63 & 41.38 & +5.75 \\
        Overall & 11,064 & 310.8 & 724.3 & 40.21 & 47.35 & +7.14 \\
        \bottomrule
    \end{tabular}
    \captionof{table}{\textbf{Complete final-input coverage groups in the independent 11,064-target evaluation.} Gains are absolute score differences (\%) computed from unrounded scores. Group membership is determined by final \texttt{CFull} visibility. This evaluation is distinct from the main training--inference crossing.}
    \label{tab:target_coverage_complete}
\end{minipage}
\end{LCAppendixBlock}

\begin{LCAppendixBlock}
\noindent
\begin{minipage}[t]{\LCAppendixColumnWidth}
\vspace{0pt}
\centering
    \small
    \setlength{\tabcolsep}{3pt}
    \setlength{\aboverulesep}{0pt}
    \setlength{\belowrulesep}{0pt}
    \setlength{\extrarowheight}{1pt}
    \renewcommand{\arraystretch}{1.0}

    \begin{tabular}{@{}lrr@{}}
        \toprule
        \rowcolor{LCHeaderGray}
        \textbf{View / conditional gain} & \textbf{Original} & \textbf{\shortstack{Without target\\reference proof}} \\
        \midrule
        \multicolumn{3}{@{}l}{\textbf{LLaDA}} \\
        \texttt{C0} & 31.0 & 31.0 \\
        \texttt{CMath} & 39.9 & 38.2 \\
        \texttt{CEnv} & 41.4 & 40.3 \\
        \texttt{CFull} & 48.5 & 45.7 \\
        $\Delta_{M\mid E}$ & +7.1 & +5.4 \\
        $\Delta_{E\mid M}$ & +8.6 & +7.5 \\
        \midrule
        \multicolumn{3}{@{}l}{\textbf{Llama}} \\
        \texttt{C0} & 32.5 & 32.5 \\
        \texttt{CMath} & 41.6 & 39.8 \\
        \texttt{CEnv} & 40.9 & 39.9 \\
        \texttt{CFull} & 49.4 & 46.3 \\
        $\Delta_{M\mid E}$ & +8.5 & +6.4 \\
        $\Delta_{E\mid M}$ & +7.8 & +6.5 \\
        \bottomrule
    \end{tabular}
    \captionof{table}{\textbf{Context reconstruction on a fixed 2,000-target subset at \texttt{CFull}-trained checkpoints.} View scores are Pass@32 (\%); conditional gains are absolute score differences (\%). Reconstruction can change both MC and EC.}
    \label{tab:reference_free_selection}
\end{minipage}\hfill
\begin{minipage}[t]{\LCAppendixColumnWidth}
\vspace{0pt}
\centering
    \small
    \setlength{\tabcolsep}{3pt}
    \setlength{\aboverulesep}{0pt}
    \setlength{\belowrulesep}{0pt}
    \setlength{\extrarowheight}{1pt}
    \renewcommand{\arraystretch}{1.0}

    \begin{tabular}{@{}lrrrr@{}}
        \toprule
        \rowcolor{LCHeaderGray}
         & \multicolumn{2}{c}{\textbf{LLaDA}} & \multicolumn{2}{c}{\textbf{Llama}} \\
        \cmidrule(lr){2-3}
        \cmidrule(lr){4-5}
        \rowcolor{LCHeaderGray}
        \textbf{Coverage group} & \textbf{Base.} & \textbf{Priority} & \textbf{Base.} & \textbf{Priority} \\
        \midrule
        Zero & 530 & 260 & 515 & 245 \\
        Partial & 580 & 350 & 570 & 340 \\
        Complete & 458 & 958 & 480 & 980 \\
        No local dependencies & 280 & 280 & 280 & 280 \\
        Unresolved & 152 & 152 & 155 & 155 \\
        \bottomrule
    \end{tabular}
    \captionof{table}{\textbf{Final target-coverage distributions on the fixed 2,000 selection-control targets.} Base. denotes the shared-pool baseline.}
    \label{tab:priority_coverage_distribution}
\end{minipage}
\end{LCAppendixBlock}

\begin{LCAppendixBlock}
\noindent
\begin{minipage}[t]{\LCAppendixColumnWidth}
\vspace{0pt}
\centering
    \small
    \setlength{\tabcolsep}{3pt}
    \setlength{\aboverulesep}{0pt}
    \setlength{\belowrulesep}{0pt}
    \setlength{\extrarowheight}{1pt}
    \renewcommand{\arraystretch}{1.0}

    \begin{tabular}{@{}lrrrrr@{}}
        \toprule
        \rowcolor{LCHeaderGray}
        \textbf{Model} & \textbf{Base.} & \textbf{Priority} & \textbf{Gained} & \textbf{Lost} & \textbf{Net} \\
        \midrule
        LLaDA & 878 & 906 & 76 & 48 & +28 \\
        Llama & 891 & 925 & 88 & 54 & +34 \\
        \bottomrule
    \end{tabular}
    \captionof{table}{\textbf{Paired solved-target counts under the shared-pool policies.} A target is solved when at least one of its 32 attempts passes Lean verification.}
    \label{tab:selector_paired_outcomes}
\end{minipage}\hfill
\begin{minipage}[t]{\LCAppendixColumnWidth}
\vspace{0pt}
\centering
    \small
    \setlength{\tabcolsep}{3pt}
    \setlength{\aboverulesep}{0pt}
    \setlength{\belowrulesep}{0pt}
    \setlength{\extrarowheight}{1pt}
    \renewcommand{\arraystretch}{1.0}

    \begin{tabular}{@{}lrrrr@{}}
        \toprule
        \rowcolor{LCHeaderGray}
         &  & \multicolumn{2}{c}{\textbf{Coverage (\%)}} &  \\
        \cmidrule(lr){3-4}
        \rowcolor{LCHeaderGray}
        \textbf{Model} & \textbf{Targets} & \textbf{Base.} & \textbf{Priority} & \textbf{Decreased} \\
        \midrule
        LLaDA & 1,120 & 24.64 & 22.41 & 170 \\
        Llama & 1,120 & 25.54 & 23.84 & 175 \\
        \bottomrule
    \end{tabular}
    \captionof{table}{\textbf{Post-selection audit of additional reference-proof dependencies.} Coverage is a macro average over auditable targets; Decreased counts targets whose coverage falls. Reference proofs do not guide either policy.}
    \label{tab:selection_tradeoffs}
\end{minipage}
\end{LCAppendixBlock}

\LCAppendixSectionGap
\linenumbers
\begin{multicols}{2}
\section{Representation Analysis Details}
\label{sec:appendix_cka}

This appendix details the reference-conditioned probe used to compare representation responses across context views and reports the complete layer-wise results.

\subsection{Shared Targets and Conditions}
\label{sec:cka_conditions}

We randomly sample 4,000 test targets with seed 42 and use the same target identities for Llama and LLaDA. For each backbone, we evaluate the four training checkpoints, \texttt{C0}, \texttt{CMath}, \texttt{CEnv}, and \texttt{CFull}, under all four inference views. For each target, the statement and reference proof remain fixed across these conditions, and input construction follows the benchmark's view-specific formatting and budget handling. This gives 16 forward conditions per backbone. Models run in evaluation mode with gradients disabled and parameters fixed.

Reference-proof conditioning aligns the proof content across views while allowing the supplied context to vary. We extract hidden states after Transformer blocks 8, 16, 24, and 32. These indices identify Transformer blocks rather than the embedding output. Token positions are defined relative to each tokenized proof, so differences in context length do not shift the proof region used for pooling.

\subsection{Model-Specific Readouts and Token Pooling}
\label{sec:cka_readouts}

For target $i$, let $L_i$ be the reference-proof length under the backbone's tokenizer. We use the central proof window
\begin{equation}
    S_i =
    \{t : \lceil 0.25L_i \rceil
    \leq t \leq \lfloor 0.75L_i \rfloor\}.
    \label{eq:cka_proof_window}
\end{equation}
The readout within this window follows each backbone's prediction setup.

For Llama, we append the reference proof to the view-specific input and perform a teacher-forced causal forward pass. To represent prediction of proof token $y_t$, we read the hidden state at its preceding causal position, denoted $h_{i,v}^{(\ell)}(t-1)$, where $v$ is the inference view and $\ell$ is the observed block. The target-level representation is
\begin{equation}
    z_{i,v}^{(\ell)}
    = \frac{1}{|S_i|}
    \sum_{t\in S_i}
    h_{i,v}^{(\ell)}(t-1).
    \label{eq:cka_llama_pooling}
\end{equation}
Causal attention restricts each prediction state to the input and preceding proof tokens.

For LLaDA, we corrupt the reference proof at a fixed masking rate of 50\% and perform one forward pass on the resulting input. Within LLaDA, a target's mask pattern is reused across all four training checkpoints and four inference views. Let $U_i$ contain the masked positions within $S_i$. We average their Transformer hidden states:
\begin{equation}
    z_{i,v}^{(\ell)}
    = \frac{1}{|U_i|}
    \sum_{t\in U_i}
    h_{i,v}^{(\ell)}(t).
    \label{eq:cka_llada_pooling}
\end{equation}
This readout characterizes a fixed corrupted-proof state. Mask positions are defined using LLaDA's tokenizer, while target identities and reference-proof text are shared across backbones.

Pooling uses the specified proof positions and excludes context, target, padding, and special-token positions. Each target contributes one vector per condition and block. We apply no additional per-example normalization, dimensionality reduction, or whitening.

\subsection{Centered Linear CKA}
\label{sec:cka_metric}
CKA normalizes a kernel dependence measure based on HSIC~\citep{gretton2005hsic,kornblith2019similarity}. For a fixed checkpoint and block, we stack the pooled vectors into one matrix per inference view, $X_v^{(\ell)}\in\mathbb{R}^{4000\times d}$, with identical target ordering across matrices. We center each feature across the $n=4000$ target rows:
\begin{equation}
    \begin{aligned}
        X_c &= X-\mathbf{1}\mu_X^\top,\\
        \mu_X &= \frac{1}{n}X^\top\mathbf{1},
    \end{aligned}
    \label{eq:cka_centering}
\end{equation}
where $\mathbf{1}\in\mathbb{R}^{n}$ is the all-ones vector; $Y$ is centered in the same way. For the centered matrices, linear CKA is~\citep{kornblith2019similarity}
\begin{equation}
    \mathrm{CKA}(X,Y)
    =
    \frac{\|X_c^\top Y_c\|_F^2}
    {\|X_c^\top X_c\|_F
     \|Y_c^\top Y_c\|_F}.
    \label{eq:cka_linear}
\end{equation}
We report dissimilarity as $D(X,Y)=1-\mathrm{CKA}(X,Y)$; larger values indicate lower representation similarity. After collecting all forward-pass batches, we align the target rows and compute CKA from the complete matrices. We retain one result per backbone, training checkpoint, observed block, and inference-view pair, recording the sample count, sampling seed, CKA, and $1-\mathrm{CKA}$.

The four inference views yield six unordered pairs within each checkpoint. Four checkpoints and four observed blocks therefore produce 96 comparisons per backbone and 192 overall. Each CKA value compares inference views within a single checkpoint; training dependence is examined by comparing these within-checkpoint contrasts across training conditions.

\subsection{Complete Layer-Wise Results}
\label{sec:cka_layerwise_results}

Figure~\ref{fig:cka_depth_mc_response} compares the
\texttt{CEnv}--\texttt{CFull} inference views across
blocks for all four training checkpoints.
Table~\ref{tab:cka_all_layers} provides the complete
results for all six inference-view pairs.

The depth patterns differ between backbones.
For the \texttt{CFull}-trained Llama checkpoint,
\texttt{CEnv}--\texttt{CFull} dissimilarity increases
from 0.0618 at block 8 to 0.1088 at block 24,
then decreases to 0.1038 at block 32.
For LLaDA, it increases from 0.0579 at block 8
to 0.1204 at block 32.
These results extend the main-text block-32
observations to earlier blocks.
\end{multicols}
\nolinenumbers

\begin{LCAppendixBlock}
\centering

    \includegraphics[width=0.94\textwidth]{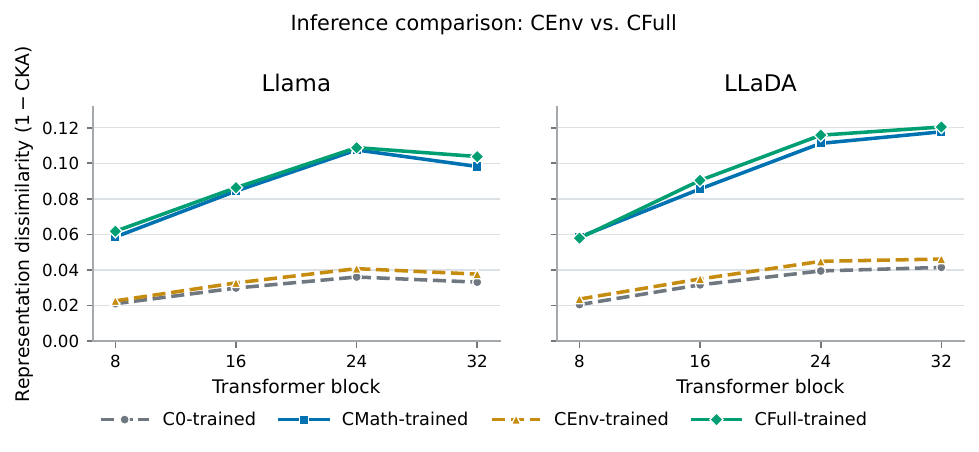}
    \captionof{figure}{\textbf{Layer-wise representation response to
    supplying MC alongside EC.} Each panel compares
    the \texttt{CEnv} and \texttt{CFull} inference
    views within each of four training checkpoints
    on the same 4,000 test targets. Points report
    $1-\mathrm{CKA}$ at blocks 8, 16, 24, and 32;
    lines connect the observed blocks.}
    \label{fig:cka_depth_mc_response}
\end{LCAppendixBlock}

\begin{LCAppendixBlock}
\centering
    \small
    \setlength{\tabcolsep}{4pt}
    \setlength{\aboverulesep}{0pt}
    \setlength{\belowrulesep}{0pt}
    \setlength{\extrarowheight}{1pt}
    \renewcommand{\arraystretch}{1.0}

    \begin{tabular}{@{}lrrrrrrr@{}}
        \toprule
        \rowcolor{LCHeaderGray}
        \textbf{\shortstack{\textbf{Training}\\\textbf{view}}} & \textbf{Block} & \textbf{\shortstack{\texttt{C0}--\\\texttt{CMath}}} & \textbf{\shortstack{\texttt{C0}--\\\texttt{CEnv}}} & \textbf{\shortstack{\texttt{C0}--\\\texttt{CFull}}} & \textbf{\shortstack{\texttt{CMath}--\\\texttt{CEnv}}} & \textbf{\shortstack{\texttt{CMath}--\\\texttt{CFull}}} & \textbf{\shortstack{\texttt{CEnv}--\\\texttt{CFull}}} \\
        \midrule
        \multicolumn{8}{@{}l}{\textbf{Llama}} \\
        \texttt{C0} & 8 & 0.0188 & 0.0234 & 0.0333 & 0.0252 & 0.0247 & 0.0211 \\
         & 16 & 0.0251 & 0.0298 & 0.0485 & 0.0376 & 0.0347 & 0.0299 \\
         & 24 & 0.0329 & 0.0402 & 0.0658 & 0.0497 & 0.0432 & 0.0361 \\
         & 32 & 0.0344 & 0.0390 & 0.0609 & 0.0475 & 0.0388 & 0.0332 \\
        \addlinespace[2pt]
        \texttt{CMath} & 8 & 0.0588 & 0.0189 & 0.0721 & 0.0630 & 0.0238 & 0.0586 \\
         & 16 & 0.0809 & 0.0273 & 0.1019 & 0.0878 & 0.0303 & 0.0844 \\
         & 24 & 0.1046 & 0.0390 & 0.1299 & 0.1167 & 0.0377 & 0.1075 \\
         & 32 & 0.0998 & 0.0363 & 0.1207 & 0.1060 & 0.0388 & 0.0982 \\
        \addlinespace[2pt]
        \texttt{CEnv} & 8 & 0.0213 & 0.0551 & 0.0725 & 0.0600 & 0.0579 & 0.0227 \\
         & 16 & 0.0323 & 0.0779 & 0.0999 & 0.0853 & 0.0775 & 0.0328 \\
         & 24 & 0.0376 & 0.1018 & 0.1324 & 0.1102 & 0.1019 & 0.0408 \\
         & 32 & 0.0395 & 0.0913 & 0.1225 & 0.1018 & 0.0953 & 0.0377 \\
        \addlinespace[2pt]
        \texttt{CFull} & 8 & 0.0589 & 0.0503 & 0.0885 & 0.0636 & 0.0551 & 0.0618 \\
         & 16 & 0.0832 & 0.0761 & 0.1287 & 0.0957 & 0.0755 & 0.0863 \\
         & 24 & 0.1034 & 0.0958 & 0.1655 & 0.1216 & 0.0961 & 0.1088 \\
         & 32 & 0.0996 & 0.0881 & 0.1558 & 0.1108 & 0.0873 & 0.1038 \\
        \midrule
        \multicolumn{8}{@{}l}{\textbf{LLaDA}} \\
        \texttt{C0} & 8 & 0.0204 & 0.0240 & 0.0369 & 0.0289 & 0.0251 & 0.0206 \\
         & 16 & 0.0312 & 0.0358 & 0.0560 & 0.0421 & 0.0370 & 0.0317 \\
         & 24 & 0.0357 & 0.0445 & 0.0693 & 0.0543 & 0.0430 & 0.0395 \\
         & 32 & 0.0398 & 0.0462 & 0.0701 & 0.0545 & 0.0483 & 0.0415 \\
        \addlinespace[2pt]
        \texttt{CMath} & 8 & 0.0540 & 0.0215 & 0.0693 & 0.0588 & 0.0206 & 0.0585 \\
         & 16 & 0.0832 & 0.0315 & 0.1042 & 0.0918 & 0.0331 & 0.0856 \\
         & 24 & 0.1074 & 0.0393 & 0.1352 & 0.1148 & 0.0428 & 0.1112 \\
         & 32 & 0.1144 & 0.0412 & 0.1403 & 0.1212 & 0.0474 & 0.1177 \\
        \addlinespace[2pt]
        \texttt{CEnv} & 8 & 0.0239 & 0.0556 & 0.0681 & 0.0609 & 0.0574 & 0.0237 \\
         & 16 & 0.0313 & 0.0806 & 0.1066 & 0.0879 & 0.0830 & 0.0350 \\
         & 24 & 0.0408 & 0.1018 & 0.1311 & 0.1157 & 0.1033 & 0.0449 \\
         & 32 & 0.0418 & 0.1061 & 0.1397 & 0.1211 & 0.1082 & 0.0462 \\
        \addlinespace[2pt]
        \texttt{CFull} & 8 & 0.0566 & 0.0520 & 0.0890 & 0.0657 & 0.0513 & 0.0579 \\
         & 16 & 0.0871 & 0.0771 & 0.1351 & 0.0963 & 0.0787 & 0.0904 \\
         & 24 & 0.1065 & 0.0978 & 0.1698 & 0.1216 & 0.0995 & 0.1158 \\
         & 32 & 0.1158 & 0.1039 & 0.1783 & 0.1312 & 0.1065 & 0.1204 \\
        \bottomrule
    \end{tabular}
    \captionof{table}{\textbf{Complete layer-wise representation dissimilarities ($1-\mathrm{CKA}$) on 4,000 shared test targets.} Training view identifies the checkpoint; each pair column compares two inference views within that checkpoint. Llama uses central-window causal prediction states, and LLaDA uses masked-position states from a reference proof corrupted at a fixed 50\% masking rate. All comparisons use the same target ordering.}
    \label{tab:cka_all_layers}
\end{LCAppendixBlock}

\LCAppendixSectionGap
\linenumbers
\begin{multicols}{2}
\section{Additional Related Work}
\label{sec:additional_related_work}

\paragraph{Premise Selection in Large Libraries.}
Selecting useful facts from a formal library has long been a central problem in automated theorem proving. Corpus-based methods use proof dependencies to learn which premises may support a new conjecture~\citep{alama2014premise}. Mining intermediate lemmas can further expand the resources available for subsequent proofs~\citep{kaliszyk2015millions}. DeepMath learns premise relevance with neural sequence models, while Magnushammer uses transformer-based retrieval and reranking~\citep{alemi2016deepmath,mikula2024magnushammer}. These approaches connect the resources made available to a prover with their usefulness in proof construction.

\paragraph{Proof Corpora and Interactive Environments.}
Formal proof corpora support learning at several levels of granularity. HOLStep supplies intermediate HOL Light statements for learning whether a statement is useful for a conjecture~\citep{kaliszyk2017holstep}. HOList provides a HOL Light environment for interacting with goals and constructing proofs~\citep{bansal2019holist}. CoqGym collects proofs from diverse Coq projects and supports tactic generation from intermediate proof states~\citep{yang2019coqgym}. These resources enable evaluation of fact selection, proof-step prediction, and interactive proof construction.

\paragraph{Formal Mathematical Benchmarks.}
Benchmarks have also expanded the mathematical problems used to evaluate provers. miniF2F aligns competition problems across proof assistants, while ProofNet pairs undergraduate-level formal statements with informal statements and proofs~\citep{zheng2022minif2f,azerbayev2023proofnet}. FIMO and PutnamBench extend evaluation to challenging Olympiad and undergraduate competition mathematics~\citep{liu2023fimo,tsoukalas2024putnambench}. FormalMATH broadens the scale and domain coverage of Lean evaluations, and CombiBench focuses on combinatorial mathematics with both proof and fill-in-the-blank tasks~\citep{yu2025formalmath,liu2025combibench}. This development makes mathematical domain, difficulty, and task format important dimensions of benchmark coverage.

\paragraph{Neural Proof Search and Automation.}
Neural provers differ in how they construct and check a proof. GPT-f generates proof steps in Metamath, and TacticToe learns tactics from HOL4 proofs to guide search~\citep{polu2020generative,gauthier2021tactictoe}. HyperTree Proof Search combines proof search with online training from search experience~\citep{lample2022hypertree}. Thor integrates language models with hammer-based automation, while Lean Copilot provides tactic suggestions, premise selection, and proof search within Lean~\citep{jiang2022thor,song2025leancopilot}. Baldur instead generates complete proofs and can use checker feedback for repair~\citep{first2023baldur}. These workflows offer different opportunities to obtain additional information during proof construction.

\paragraph{Proof Data and Prover Training.}
Several approaches improve proving by changing the supervision or pretraining data. PACT derives auxiliary training tasks from kernel-level proof artifacts~\citep{han2022pact}. Llemma continues pretraining on mathematical text and code, supporting formal proving without task-specific fine-tuning~\citep{azerbayev2024llemma}. LEAN-GitHub extracts proofs and tactic states from repositories beyond Mathlib, while the original DeepSeek-Prover builds synthetic Lean statements and proofs for training~\citep{wu2024leangithub,xin2024deepseekprover}. STP jointly trains a conjecturer and a prover through iterative conjecturing and verified proof generation~\citep{dong2025stp}. These methods show that prover training can draw on human-written proofs, proof artifacts, and generated mathematical problems.

\paragraph{Informal Reasoning and Formal Proof Guidance.}
Informal mathematical information provides another source of guidance. Autoformalization translates natural-language mathematical statements into formal specifications~\citep{wu2022autoformalization}. Draft, Sketch, and Prove converts informal proofs into formal sketches and uses automated proving to complete the resulting subproblems~\citep{jiang2023draft}. Lean-STaR trains models to generate informal thoughts before formal proof steps, using synthetic thoughts paired with tactics and subsequent expert iteration~\citep{lin2025leanstar}. Here, the added guidance takes the form of translated specifications, proof sketches, or reasoning annotations.

\paragraph{Reusable Declarations and Evolving Libraries.}
The mathematical resources available to a prover can change as a project develops. LEGO-Prover retrieves verified lemmas from a growing library and adds reusable lemmas produced during proof construction~\citep{wang2024legoprover}. Graph2Tac builds representations for newly encountered definitions, allowing a prover to use concepts beyond its original training setting~\citep{blaauwbroek2024graph2tac}. LeanAgent combines a dynamic database with progressive training across Lean repositories~\citep{kumarappan2025leanagent}. These approaches connect access to reusable mathematical resources with the model's ability to work in an evolving formal library.

\paragraph{Representation Comparison.}
Representation similarity offers a complementary way to study model responses. SVCCA combines dimensionality reduction with canonical correlation analysis, and projection-weighted CCA weights the resulting correlations when comparing representations~\citep{raghu2017svcca,morcos2018pwcca}. CKA normalizes a kernel dependence measure related to HSIC and supports comparison of neural representations~\citep{gretton2005hsic,kornblith2019similarity}. In \textsc{LeanContext}, linear CKA compares input views under fixed reference-proof conditioning. The resulting responses complement the verified-success and matched-content comparisons, which assess the observable benefits of supplied declarations.
\end{multicols}
\nolinenumbers